%% file: root.tex
\documentclass[letterpaper, 10 pt, conference]{ieeeconf}  

\IEEEoverridecommandlockouts                              
\let\labelindent\relax
\usepackage[shortlabels]{enumitem}
\input{macros_latex}  
\input{macros_math}   

\usepackage{pifont}
\newcommand{\cmark}{\color{green}\ding{51}}%
\newcommand{\pcmark}{\color{blue}\rotatebox[origin=c]{90}{\ding{121}}}%
\newcommand{\xmark}{\color{red}\ding{55}}%

\graphicspath{{figures/}}

\usepackage[ruled,vlined]{algorithm2e}
\LinesNumbered
\DontPrintSemicolon
\SetNoFillComment 
\SetCommentSty{mycommfont}
\SetKwComment{Comment}{$\triangleright$\ }{}
\SetAlCapFnt{\footnotesize} 
\SetAlFnt{\footnotesize} 
\SetNlSty{textnormal}{\scriptsize}{} 

\if\isOverleaf%
    
\else
    
\fi
\usepackage[frozencache=true, cachedir=minted-cache, newfloat]{minted}
\setmintedinline[python]{fontsize=\normalsize, gobble=0}
\usepackage{tcolorbox}
\tcbuselibrary{listings, minted, skins}
\tcbset{listing engine=minted}
\definecolor{bg}{RGB}{255, 255, 255}
\newtcblisting{pythonlst}{listing only, minted language=python, minted style=emacs, colback=bg, enhanced, frame hidden, minted options={fontsize=\fontsize{6.75pt}{6.75pt}\selectfont, tabsize=2, autogobble, linenos}}
\newenvironment{python}{\captionsetup{type=listing}}{}
\SetupFloatingEnvironment{listing}{name=Code-Example}

\def\vspaceabove{0em}  
\def\vspacebelow{0em}  

\definecolor{revisions}{rgb}{0, 0, 0}
\definecolor{black}{rgb}{0, 0, 0}
\newcommand{\rev}[1]{\color{revisions}#1\color{black}}

\title{EAGERx: Graph-Based Framework for Sim2real Robot Learning}
\usepackage{times}

\author{Bas van der Heijden$^{*1}$, Jelle Luijkx$^{*1}$, Laura Ferranti$^{1}$, Jens Kober$^{1}$, Robert Babuska$^{1}$
\thanks{$^{*}$Equal contribution}%
\thanks{$^{1}$ Dept. of Cognitive  Robotics, Delft  University  of  Technology, The Netherlands. {\tt\small d.s.vanderheijden@tudelft.nl}}%
\thanks{This work is funded by the EU's H2020 OpenDR project (grant No 871449) and the Dutch Science Foundation NWO-TTW's Veni project HARMONIA (18165).}%
}

\begin{document}

\maketitle

\begin{abstract}%
	Sim2real, that is, the transfer of learned control policies from simulation to real world, is an area of growing interest in robotics due to its potential to efficiently handle complex tasks. 
    The sim2real approach faces challenges due to mismatches between simulation and reality. 
    These discrepancies arise from inaccuracies in modeling physical phenomena and asynchronous control, among other factors.
    To this end, we introduce EAGERx, a framework with a unified software pipeline for both real and simulated robot learning.
    It can support various simulators and aids in integrating state, action and time-scale abstractions to facilitate learning.
	EAGERx's integrated delay simulation, domain randomization features, and proposed synchronization algorithm contribute to narrowing the sim2real gap.
	We demonstrate (in the context of robot learning and beyond) the efficacy of EAGERx in accommodating diverse robotic systems and maintaining consistent simulation behavior. EAGERx is open source and its code is available at \url{https://eagerx.readthedocs.io}.
\end{abstract}


\input{introduction.tex}
\input{framework.tex}
\input{synchronization.tex}
\input{application.tex}
\input{discussion.tex}
\input{conclusion}

\bibliographystyle{IEEEtran}

\bibliography{glorified, new}  

\end{document}

%% file: macros_latex.tex
\usepackage[table]{xcolor}  
\usepackage{amsmath}
\usepackage{amssymb}
\usepackage{graphicx}
\usepackage{url}
\usepackage{pgf}
\usepackage{subcaption}
\usepackage{float}
\usepackage{outlines}  

\usepackage[font=small]{caption}

\def\secref#1{Sec.~\ref{#1}}
\def\figref#1{Fig.~\ref{#1}}
\def\tabref#1{Tab.~\ref{#1}}
\def\eqref#1{Eq.~(\ref{#1})}
\def\algref#1{Alg.~\ref{#1}}
\def\coderef#1{Code-Example~\ref{#1}}

\makeatletter
\usepackage{xspace}
\DeclareRobustCommand\onedot{\futurelet\@let@token\@onedot}
\def\@onedot{\ifx\@let@token.\else.\null\fi\xspace}

\makeatother

\def\citep#1{\cite{#1}}

\def\StripPrefix#1>{}
\def\isOverleaf{\fi
    \def\overleafJobname{output}
    \edef\overleafJobname{\expandafter\StripPrefix\meaning\overleafJobname}%
    \edef\job{\jobname}%
    \ifx\job\overleafJobname
}

\usepackage[colorinlistoftodos, textsize=small, textwidth=50mm]{todonotes}
\definecolor{bhcolor}{rgb}{.7, 0.7, 0.1}
\definecolor{lfcolor}{rgb}{.7, 0.1, 0.7}
\definecolor{jlcolor}{rgb}{.1, 0.7, 0.7}
\definecolor{todocolor}{rgb}{1.0, 0.0, 0.0}

%% file: macros_math.tex
\newcommand{\SetInput}{\ensuremath{\mathcal{U}}}
\newcommand{\SetOutput}{\ensuremath{\mathcal{Y}}}

\newcommand{\delay}{\ensuremath{\tau}}
\newcommand{\rate}{\ensuremath{f}}
\newcommand{\dt}{\ensuremath{\Delta t}}

\newcommand{\floor}[1]{\ensuremath{\lfloor{#1}\rfloor}}

%% file: introduction.tex
\section{Introduction}
\label{sec:intro}
\begin{figure}[t]
    \centering
    \includegraphics[width=\columnwidth]{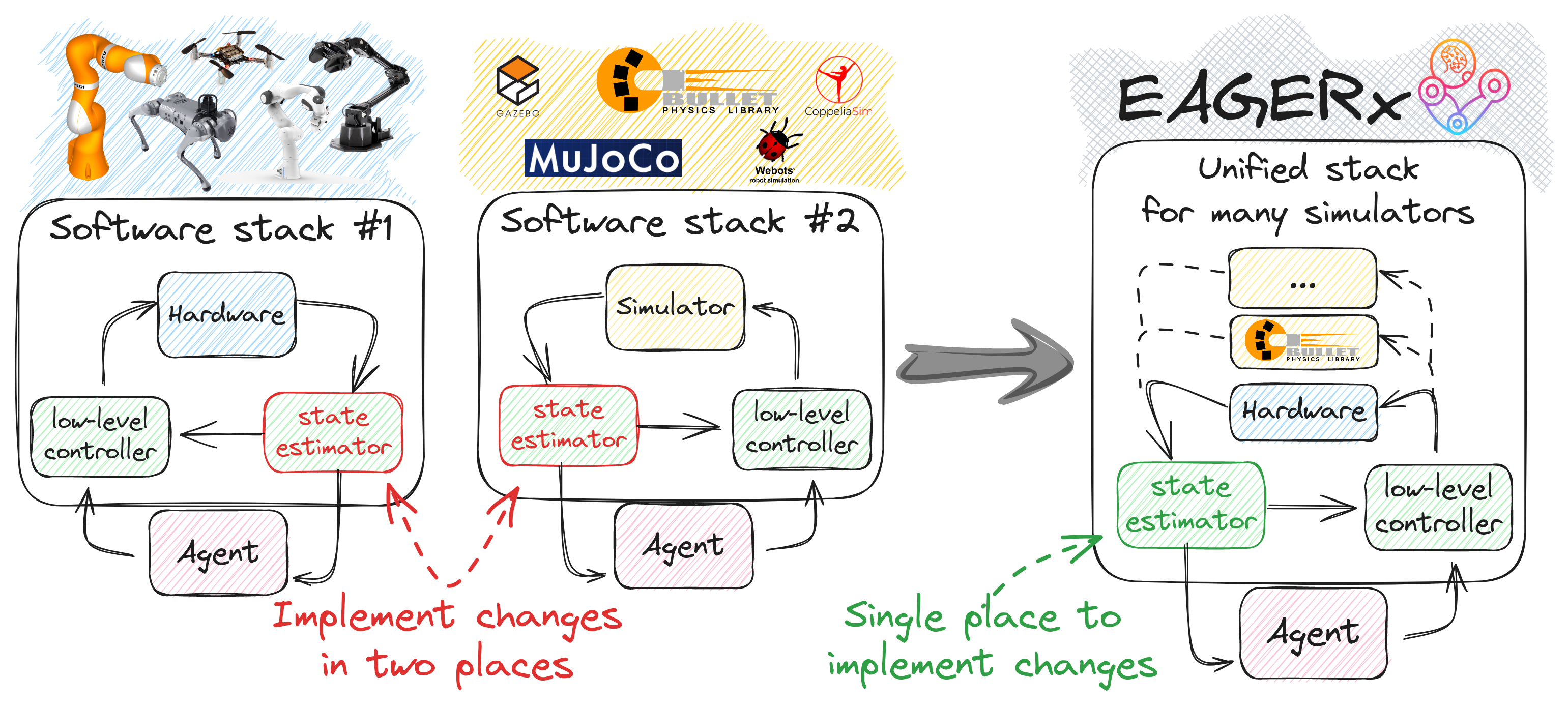}
    \caption{
    Our framework offers a unified software pipeline for both simulated and real robot learning. 
    It can support various simulators and aids in integrating state, action and time-scale abstractions.
    }
    \vspace{\vspacebelow}
    \label{fig:framework}
\end{figure}

Transferring control policies trained in simulation to the real world, known as sim2real, has gained considerable interest in the field of robotics due to its potential to address complex tasks with remarkable efficiency \citep{rudin2022learning, tan2018sim}.  
Simulations offer a safe, cost-effective, and controlled environment for training and testing robotic algorithms, allowing roboticists to refine their models and controllers without the risks and expenses associated with real-world experimentation.
The sim2real approach, however, faces challenges due to the \emph{sim2real gap}, that is, unaccounted discrepancies between simulation and reality. 
These disparities may stem from \emph{inaccurate modeling of physical phenomena} (e.g.,  friction, deformations, and collisions) or from the use of \emph{separate software implementations} for reality and simulation, which may lead to unintended mismatches as depicted in \figref{fig:framework}.
Another subtle but significant source of discrepancy is the \emph{asynchronous nature} of robotic systems. 
While robotic systems are typically simulated sequentially \cite{brockman2016openai}, sensing, computation, and acting happen concurrently in reality.
Disregarding these differences can be detrimental to the real-world performance of a policy trained in simulation.\looseness=-1


Inaccurate modeling of physical phenomena in simulation is typically mitigated by domain randomization \citep{tan2018sim}.
However, this approach can make the simulation more challenging, which may lead to longer training times and suboptimal policies. 
Reformulating the task to the right level of abstraction may be more effective to alleviate the sim2real gap if the abstraction captures the task and can be extracted accurately both from simulated and real data \citep{hofer2021sim2real}.
Abstractions can take various forms, such as action abstraction that simplifies control issues using high-level actions, time-scale abstraction that uses macro-actions for multi-scale planning and learning, and state abstraction that condenses raw sensor data into key features \citep{hofer2021sim2real}. 
Therefore, existing sim2real frameworks \cite{lucchi2020robo, gonzalez2019gym} have exploited the multi-rate graph-based design of ROS \cite{quigley2009ros} to obtain a unified software pipeline that allows for the integration of various kinds of abstractions.  
However, these frameworks restrict users to the Gazebo simulator \cite{koenig2004design}, which can be limiting as different tasks may require specific types of simulators.
Additionally, these frameworks fall short in synchronizing components that operate in parallel within the simulation.
At faster-than-real-time simulation speeds, this can exacerbate communication and processing delays, leading to inconsistencies, inaccuracies, and potential system instability. 
Such amplified delays can compromise the proper functioning of the simulated system, rendering learned policies ineffective when transferred to real-world environments. 
Conversely, naive synchronization may also widen the sim2real gap if it overlooks the concurrent nature of sensing, computation, and acting in reality.

\rev{In addition to ROS-based frameworks, existing robot learning frameworks provide integration of abstractions through a modular design and unified framework, often coupled with a specific simulator.
Notable examples include Isaac Orbit \cite{mittal2023orbit} and Drake \cite{tedrake2019drake}.
Isaac Orbit is a modular robot learning framework built on top of the Isaac Sim simulator \cite{nvidia2022isaacsim}, offering benchmarks and readily available robot models for convenient experimentation.
On the other hand, Drake is a model-based framework combining a multibody dynamics engine with a systems approach and optimization framework \cite{tedrake2019drake}.
However, these frameworks are tied to a single simulator, while various robot simulators are available, each with its own strengths and weaknesses.
Existing robot learning frameworks lack the flexibility to choose a simulator or leverage various simulators' strengths.
}

The main contribution of this paper is EAGERx (Engine Agnostic Graph Environments for Robotics), that is, a robot learning framework with a unified software pipeline compatible with both simulated and real robots that supports the integration of various abstractions and simulators as depicted in \figref{fig:framework}.
EAGERx introduces a novel synchronization protocol that coordinates inter-node communication based on node rates and anticipated delays.
By simulating delays, our protocol maintains asynchronous robotic system relationships \emph{synchronously}, preserving the benefits of modular and synchronous simulation.
Contrasting with sequential simulation, the protocol permits nodes to transmit messages asynchronously and perform tasks without waiting for immediate responses, thereby accelerating the simulation and allowing nodes to progress based on their processing capabilities and data availability.
EAGERx is Python-based and offers high simulation accuracy without compromising speed, native support for domain randomization and delay simulation, and a modular structure for easy manual reset procedures and prior knowledge integration.
Our framework features a consistent interface, an interactive GUI, \rev{continuous integration with tests covering 94\% of the code}, and comprehensive documentation, including interactive tutorials, easing new user adoption. The documentation, tutorials, and our open-source code can be found at \url{https://eagerx.readthedocs.io}.

In summary, we make four key contributions:
\begin{enumerate}[start=1,label={\bfseries C\arabic*}]
    \item A synchronization protocol that ensures consistent simulation behavior even beyond real-time speeds.
    \item A modular design that \rev{can } support various robotic systems and state, action, and time-scale abstractions.
    \item An agnostic design that allows compatibility with multiple engines.
    \item \rev{The }integrated delay simulation and domain randomization \rev{features } in EAGERx can narrow the sim2real gap.
\end{enumerate}
\rev{
    The remainder of the paper is structured as follows.
    \secref{sec:framework} provides a high-level introduction to the framework.
    \secref{sec:synchronization} elaborates on a key low-level component of the framework, i.e., its novel synchronization algorithm.
    \secref{sec:experimental_evaluation} provides an extensive experimental evaluation to show the applicability of EAGERx for sim2real robot learning.
    \secref{sec:beyond_rl} shows the framework's utility beyond sim2real robot learning in two real-world robotic use cases.
    \secref{sec:discussion} compares EAGERx and existing frameworks, and \secref{sec:conclusion} concludes the paper.
    }

%% file: framework.tex
\section{Framework}
\label{sec:framework}
This section provides an overview of EAGERx. 
\secref{sec:agnostic} outlines the framework's main components. Then, \secref{sec:support} discusses the package management system promoting modularity and versioned compatibility. Finally, \secref{sec:sim2real} discusses the framework's capabilities for domain randomization, simulator augmentation, and delay simulation, which are essential to minimize the sim2real gap.

\subsection{Agnostic Framework}
\label{sec:agnostic}
First, we provide a brief overview of the main components, followed by a code example.

\subsubsection{Graph}
EAGERx processes are represented as nodes within a graph structure, linked by directed edges from a node's output to one or multiple node inputs.
Nodes communicate via edges by exchanging messages.
This versatile decentralized architecture, ideal for networked hardware and off-board computer interactions, is especially useful for robotics.

\subsubsection{Node}
Nodes are central to EAGERx, representing individual processes that execute concurrently. 
Each node begins a new episode with a user-defined reset that sets its initial state, followed by the execution of user-defined code, termed a \textit{callback}, at a predetermined rate. 
These callbacks determine the node's functionality and define how inputs from other nodes are transformed into outputs that are, in turn, sent as output to subsequent nodes.
A typical robotic system usually consists of many such interconnected nodes.
For instance, one node may be responsible for capturing camera images, another for localization using these images, and yet another for directing the robot's movement based on the localization data.

Nodes can be launched in various ways, according to their operational needs. 
For example, \emph{CPU-bound nodes}, which are computationally intensive, benefit from being launched as subprocesses. 
This approach leverages multi-processing to bypass the limitations imposed by Python's Global Interpreter Lock (GIL), thus enhancing computational efficiency. 
In contrast, \emph{I/O-bound nodes}, which primarily handle input/output operations, are more efficiently launched as separate threads. 
This minimizes the overhead associated with message serialization, streamlining communication.
Furthermore, EAGERx facilitates \textit{distributed computing} by enabling nodes to be launched as external processes on different machines. 
This feature allows for the distribution of computational loads across a network, optimizing the overall performance of the robotic control system.

\subsubsection{Object}
EAGERx objects enable flexible node replacement when transitioning a robotic system from simulation to reality. 
For instance, in reality, nodes for extracting sensor data from a physics engine become obsolete, requiring replacement with nodes interfacing robot hardware. 
EAGERx objects accommodate this adaptability.

\begin{figure}[tb]
    \centering
    \begin{subfigure}[b]{0.99\columnwidth}
        \centering
        \includegraphics[width=\textwidth]{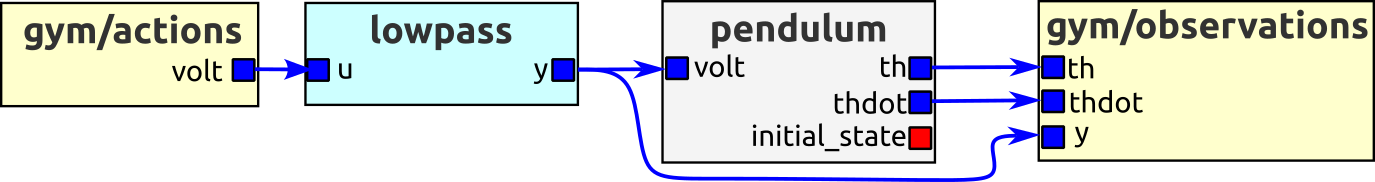}
        \caption{Agnostic graph}
        \label{fig:graph_agnostic}
    \end{subfigure}
     \par\smallskip
    \begin{subfigure}[b]{0.4\columnwidth} 
        \centering
        \includegraphics[width=\columnwidth]{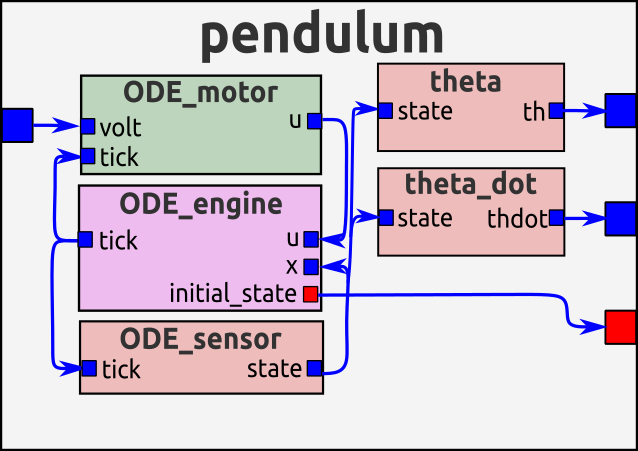}
        \caption{ODE subgraph}
        \label{fig:graph_ode}
    \end{subfigure}
    \hfill
    \begin{subfigure}[b]{0.4\columnwidth} 
        \centering
        \includegraphics[width=\columnwidth]{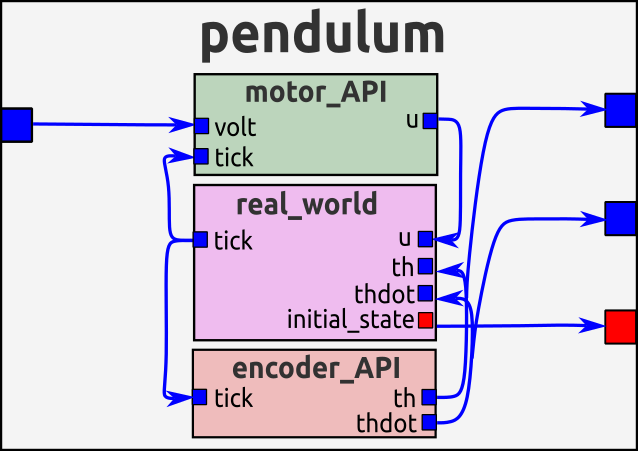}
        \caption{Real-world subgraph}
        \label{fig:graph_real}
    \end{subfigure}
    \caption{(a) Displays the engine-agnostic graph of the pendulum environment from \coderef{code:main} as generated by the GUI.
    The engine-specific subgraphs for replacing the object (i.e., pendulum) are depicted for the ODE (b) and real-world (c) engines.
    The yellow nodes, split for visualization clarity, symbolize the agent's actions and observations.
    Blue squares represent I/O channels, while red squares indicate node states and/or parameters that can be randomized at the start of an episode.
    }
    \vspace{\vspacebelow}
    \label{fig:graph}
\end{figure}

Objects define abstract inputs and outputs, as well as subgraphs for each supported physics engine.
Users can add objects to graphs (\figref{fig:graph_agnostic}), and establish connections between nodes and objects.
Upon selecting a physics engine, abstract objects are replaced by corresponding subgraphs (\figref{fig:graph_ode}, \figref{fig:graph_real}), rendering the node and object graph \textit{engine-agnostic} (\figref{fig:graph_agnostic}), as it supports multiple physics-engines.
Notice how the framework treats reality as just another physics engine. 
Practically, objects represent entities interacting directly with the physical environment.
For instance, a robot may have an abstract input and output for its motors and encoders, respectively.
Depending on the chosen physics engine, the robot's subgraph comprises nodes interfacing with real hardware or nodes communicating with a simulator.

The Object's design also accommodates the difference in available data between simulators and real-world hardware. 
They enable the definition of simulation-specific outputs, such as data exclusive to simulators, and inputs like randomized external disturbances, that can be used to enhance policy robustness. 
Users can easily configure these elements, selecting or deselecting them as needed, to ensure compatibility across different physics engines, thereby adapting the node and object graph for diverse simulation and real-world scenarios.

\subsubsection{Engine}
Physics-engines (e.g., PyBullet \citep{coumans2021pybullet}, Gazebo \citep{koenig2004design}) are interfaced by a special node called the \emph{engine}.
The engine initiates the physics engine, adds 3D meshes, and sets dynamic parameters (e.g., friction coefficients).
It controls time passage and its rate defines the simulation step size.

\subsubsection{Backend}
Node processes, launched in various ways (i.e., subprocess, multi-threaded, distributed), communicate through edges and interact with a collective database called the parameter server.
The backend facilitates low-level node-to-node communication (i.e., establishing connections and the serialization of messages) for every edge and controls the parameter server.
EAGERx supports two backends (i.e., ROS1, SingleProcess), with an abstract backend API allowing users to implement custom backends.
Defined graphs can be initialized as distributed networks of subprocesses or run in a single process.
EAGERx provides an abstraction layer over ROS, adding key features for robot learning such as synchronized faster-than-real-time simulation, domain randomization, and delay simulation.

\subsubsection{BaseEnv}
EAGERx favors composition over inheritance as a design principle because robotic systems are more naturally constructed from various components than by finding commonalities and using inheritance. 
EAGERx environments consist of an engine, backend, and graph, which is composed of nodes and objects.
This design promotes code reuse and handles future requirement changes better than an inheritance-based environment. 
Nodes operating within the graph of an EAGERx environment support multi-processing, thus enabling efficient parallel operations. 
Additionally, EAGERx facilitates vectorization across multiple environments, thereby enhancing the system's scalability and performance capabilities.
BaseEnv conforms to the OpenAI Gym interface \citep{brockman2016openai}. 
The reset method initializes episodes by setting the aggregate initial state of all graph nodes, enabling domain randomization over any registered node state, and returns the first observation.
Users then determine actions, which are relayed to connected nodes through the \textit{step} method.

\begin{python}
\begin{pythonlst}
from .tutorials.pendulum import Pendulum # Make object
o = Pendulum.make(name="pendulum")
from .tutorials.low_pass import LowPass # Make node
n = LowPass.make(name="lowpass", rate=15, cutoff=7)

from eagerx import Graph # Make `agnostic` graph
g = Graph.create([o, n])
g.connect(action="volt",      target=n.inputs.u)
g.connect(source=n.outputs.y, target=o.actuators.volt,
          delay=0.1)  # Simulates actuator delay 
g.connect(source=n.outputs.y,     observation="y",
          skip=True) # Resolves cyclic dependency
g.connect(source=o.sensors.th,    observation="th",
          window=2)  # Use last 2 sensor readings
g.connect(source=o.sensors.thdot, observation="thdot",
          window=2)  # Use last 2 sensor readings

from eagerx_ode.engine import OdeEngine # Select engine
e = OdeEngine.make(rate=30, 
                   real_time_factor=0, # 0 -> unlimited
                   sync=True) # toggles synchronization
from eagerx.backends.single_process import SingleProcess
b = SingleProcess.make()  # Make backend

from .tutorials.env import CustomEnv # Make env
env = CustomEnv(g, ode, b, name="env_id", rate=30)

obs, info = env.reset()          # Start a new episode
a    = env.action_space.sample()    # Select an action
obs, reward, terminated, truncated, info = env.step(a)
env.shutdown() # Release resources
\end{pythonlst}
\vspace{\vspaceabove}
\caption{Environment creation for the swing-up problem. \\}
\label{code:main}
\end{python}

\coderef{code:main} showcases the steps to create an environment using EAGERx for the pendulum swing-up problem, a classic problem in reinforcement learning \cite{brockman2016openai}. 
It begins with the creation of a \textit{pendulum} object and a \textit{lowpass} node to filter the agent's actions, thereby reducing wear and tear on the system (lines 1-4).
Subsequently, an \textit{agnostic} graph is constructed in which the various components are connected, anticipated delays are specified for simulation, and cyclical connections are handled (lines 6-16).
The environment is set up with the \textit{OdeEngine} physics engine and a \textit{SingleProcess} backend (lines 18-23). Equally, the \textit{RealEngine} could be used to switch to real-world scenarios.
Following initialization, an interaction is implemented by sampling an action and applying it to the environment (lines 25-30), with the environment being cleanly shut down at the end (line 31).

\subsection{Support}
\label{sec:support}
Robotic system design often involves multiple cycles of design, implementation, evaluation, and refinement. EAGERx supports the users as follows:
\subsubsection{Visualization Tools}
EAGERx offers interactive visualization tools that aid in understanding and debugging robotic systems. 
Users can visualize the graph of nodes and inspect the parameter specifications of individual nodes with EAGERx's interactive GUI.
The ability to visualize a complex robotic system is a powerful tool for debugging and understanding the system's behavior.
Example visualizations of the GUI are shown in Figures \ref{fig:graph_agnostic}, \ref{fig:pendulum_graph}, and \ref{fig:box_push_graph}.

\subsubsection{Package Management}
EAGERx incorporates a package management system that fosters modularity, versioned compatibility, and automated unit tests covering 94\% of the code.
This system allows users to easily share, import, and reuse code modules in different projects. 
By promoting modular design, EAGERx enables users to build complex robotic systems by combining smaller, well-tested components.

\subsubsection{Onboarding Resources}
EAGERx provides comprehensive onboarding resources, including interactive tutorials, code samples, and documentation, to help users quickly learn and adopt the framework.

\subsection{Mitigating the Sim2Real gap}
\label{sec:sim2real}
To address the sim2real gap, EAGERx's modular design enables manual reset routines, simulator augmentation, and supports domain randomization and delay simulation.

While resetting simulations is straightforward, real-world resets demand meticulously crafted routines to revert the system to its initial state. 
To this end, specialized reset nodes can be integrated into the graph, simplifying the real-world reset process between episodes. 
These nodes might execute procedures requiring human interaction or engage safety filters within the graph. 
Operational only during reset phases, they remain inactive during regular episodes.

Simulator augmentation in EAGERx enables the integration of custom models, capturing complex dynamics absent in standard simulations. 
For instance, in \citep{rudin2022learning}, augmenting the simulator with a custom actuator model was key to successful sim2real transfer. 
This flexibility in EAGERx enhances simulation accuracy and fidelity, thus facilitating a more effective sim2real transition.

EAGERx enables domain randomization by varying simulation parameters such as object shapes and lighting \citep{tan2018sim}. 
Within this framework, nodes can register any parameter as a state, enabling its randomization via the \textit{reset} method of the environment.

Delay simulation is enabled by our synchronization protocol discussed in \ref{sec:support}, and emulates communication latency and computational delays encountered in real-world systems, yielding a more accurate simulation.
Delays can be implemented across any graph edge, encompassing edges between nodes and objects, thus simulating sensor and actuator delays as demonstrated in \coderef{code:main}, line 10.

%% file: synchronization.tex
\section{Synchronization}
\label{sec:synchronization}
Parallel computation, used in robotic system simulations via ROS \cite{quigley2009ros} in existing sim2real frameworks \citep{lucchi2020robo, gonzalez2019gym}, can increase simulation speeds. 
When run at faster-than-real-time speeds, however, these frameworks suffer from unsynchronized parallel components, unintentionally widening the sim2real gap. 
Here, the individual computation delays become more pronounced relative to the accelerated simulation clock. 
Without suitable synchronization at high speeds, certain components may struggle to match pace and gradually fall out of sync, leading to a deviation in the simulation from its real-world counterpart.
Consequently, the learned control policy's performance may deteriorate, as it could receive outdated or mismatched observations, yielding actions based on inaccurate data.
This may render the learned policies ineffective when transferred to the real-world environment.

\subsection{Protocol}
We developed a synchronization protocol for each of the nodes representing the robotic system that enables parallel computation and minimizes additional message-passing overhead, thereby enhancing system efficiency and accuracy. 
This protocol ensures that each node's \rev{callback}, a user-defined code block executed at a predetermined rate for processing inputs and generating outputs, is triggered under the right conditions. 
Properly constructed communication patterns and protocols can achieve global synchronization without a central coordinator, whereby each node proceeds with its tasks once necessary input data or conditions have been satisfied.

Each node is launched as a subprocess that runs a local protocol version, depicted in \algref{alg:protocol}.
The conditions for a node to proceed with the next callback are based on the expected ordering of events, as dictated by assumed rates and delays of the system (lines 5-9).
Executed with an event loop thread and dedicated input channel threads, the protocol compares received and expected message counts for input channels before executing subsequent callbacks (lines 10-13).
This comparison informs whether a node proceeds with the next callback or awaits more messages.
Nodes perform tasks based on the protocol's decision and asynchronously transmit output to connected nodes (line 14).
Only upon completion of the previous callback or receipt of a new input channel message does the event-driven protocol evaluate conditions for the subsequent callback (line 17).
Consequently, task execution is entirely independent of any global clock or synchronization messages, thus minimizing additional message-passing overhead.

The protocol computes expected messages per input channel with node $n$ executing its callback at rate $\rate_n$ and receiving messages at rate $\rate_i$ delayed by $\delay_i$ over input channels $i\in\SetInput$ as summarized by \algref{alg:acyclic}.
Assuming nodes maintain their rates, callbacks occur every $\dt_n=\frac{1}{\rate_n}$ seconds, and messages are received every $\dt_i=\frac{1}{\rate_i}$ seconds.
The protocol expects the $k$th callback after $k\dt_n$ seconds, anticipating $\floor{(k\dt_n - \delay_i)/\dt_i}$ messages from each input channel $i$, where $\floor{a/b}$ denotes the integer division operator.
While this intuition underpins the synchronization protocol, the implementation in \algref{alg:acyclic} is more complex.
Computations are recast in rates to improve numerical stability by minimizing floating-point imprecision in case of high rates (small time intervals).
The protocol sets every input channel's initial expected message count to $1$, irrespective of $\delay_i$, simplifying callback implementations.

The protocol also handles the special case of cyclical dependencies-—common in robotics systems interacting with a physic-engine and can cause deadlocks otherwise—with \algref{alg:cyclic}. 
In EAGERx, users can designate input channels as cyclical, postponing dependency to the next callback. 
This strategy allows one node to execute first in a cycle, while others await this node's output.


\subsection{Limitations}
The protocol's limitations should be considered in the context of the underlying communication protocol, which must ensure the preservation of message order and be lossless. 
The protocol assumes that the robotic system can be represented by nodes with fixed rates and at least one input.
Although the protocol can be easily toggled between synchronous and asynchronous modes, it does not allow for a hybrid mode, where some nodes are synchronized and others are not.
Finally, the protocol does not account for jitter and assumes deterministic delay; however, this limitation can be mitigated by varying the delay across episodes if needed.

\begin{algorithm}[t]
	\DontPrintSemicolon
	\KwIn{node rate $\rate_n$, input rates $\rate_i$, input delay $\delay_i$, input channels $i\in\SetInput$, output channels $j\in\SetOutput$}
	\KwOut{Processed data sent to downstream nodes}
	
	\SetKwBlock{eventLoopThread}{eventLoopThread:}{}
	\SetKwBlock{inputChannelThread}{inputChannelThread $\bf{i}$:}{}
    
	$k \leftarrow$ Initialize callback index to $0$\;
    $B_i \leftarrow$ Initialize empty buffers for every input channel $i$

	Start eventLoopThread\;
	Start inputChannelThread for every $i\in\SetInput$\;
	
	\eventLoopThread{
		\ForEach{$i\in\SetInput$}{
			\If{channel $i$ is cyclical }{
				$\delta_i \leftarrow$ Expected message count (Alg.~\ref{alg:cyclic}) \;
			}
			\Else{
				$\delta_i \leftarrow$ Expected message count (Alg.~\ref{alg:acyclic}) \;

			}
		}
		\If{$\delta_i \leq \text{size}(B_i)$ for every $i\in\SetInput$}{
			\ForEach{$i\in\SetInput$}{
				$u_{i, k} \leftarrow $ Pop last $\delta_i$ messages from $B_i$
			}
			$y_k \leftarrow$ Run callback with inputs $u_{i, k}$, $\forall i\in\SetInput$\;
			Send $y_k$ to all output channels $j\in\SetOutput$ \;
			$k \leftarrow$ Increment callback index to $k + 1$\;
			Trigger event on eventLoopThread\;
		}
		WaitForEvent\;
		}
		
	\inputChannelThread{
		$B_i \leftarrow$ Buffer received message\;
		Trigger event on eventLoopThread\;
	}
	
	\caption{\footnotesize Synchronization protocol executed by each node}
	\label{alg:protocol}
\end{algorithm}
\begin{algorithm}[t]
	\DontPrintSemicolon
	\KwIn{callback index $k$, node rate $\rate_n$, input rate $\rate_i$, input delay $\delay_i$}
	\KwOut{Expected number of messages $\delta$ to receive between the $k-1$th and $k$th callback}
	
	\uIf{$k = 0$}{
		$\delta \leftarrow 1$ \tcp*{Set initial count to $1$}
	}
	\Else{
		\tcc{Expected count between $k-1$ and $k$}
		$N_{k-1} \leftarrow \floor{(\rate_i(k-1) - \rate_n\rate_i\delay_i)/\rate_n}$ \;
		$N_k \leftarrow \floor{(\rate_i(k) - \rate_n\rate_i\delay_i)/\rate_n}$ \;
		
		$\Delta \leftarrow N_k - N_{k-1}$ \;

		\tcc{Correct expected count with delay}
		$c \leftarrow \floor{(\rate_i k - \rate_n\Delta - \rate_n\rate_i\delay_i)/\rate_n}$ \;

		$\delta \leftarrow \Delta - \min(\Delta, \max(0, -c))$ \;
	}
	\caption{\footnotesize Expected number of messages to receive between the $k-1$th and $k$th callback}
	\label{alg:acyclic}
\end{algorithm}
\begin{algorithm}[t]
	\DontPrintSemicolon
	\KwIn{callback index $k$, node rate $\rate_n$, input rate $\rate_i$, input delay $\delay_i$, fudge factor $\epsilon\approx10^{-9}$}
	\KwOut{Expected number of messages $\delta$ to receive between the $k-1$th and $k$th callback}
	
	\uIf{$k = 0$}{
		$\delta \leftarrow 0$ \tcp*{Set initial count to $0$}
	}
	\Else{
		\tcc{Calculate count as if $k$ is shifted}
		\uIf{$\rate_n > \rate_i$}{
			$o \leftarrow \floor{(\rate_n - \epsilon)/\rate_i}$ \tcp*{Forward}
		}
		\Else{
			$o \leftarrow -1$ \tcp*{Backward}
		}
		
		\tcc{Expected count between $k-1$ and $k$}
		$N_{k-1} \leftarrow \floor{(\rate_i(k-1+o) - \rate_n\rate_i\delay_i)/\rate_n}$ \;
		$N_k \leftarrow \floor{(\rate_i(k+o) - \rate_n\rate_i\delay_i)/\rate_n}$ \;
		$\Delta \leftarrow N_k - N_{k-1}$ \;

		\tcc{Correct expected count with delay}
		$c \leftarrow \floor{(\rate_i k - \rate_n(\Delta - 1)- \rate_n\rate_i\delay_i)/\rate_n}$ \;
		$\delta \leftarrow \Delta - \min(\Delta, \max(0, -c))$ \;
	}
	\caption{\footnotesize Expected number of messages to receive between the $k-1$th and $k$th callback to resolve a cyclical dependency}
	\label{alg:cyclic}
\end{algorithm}

%% file: application.tex
\section{Experimental Evaluation}
\label{sec:experimental_evaluation}

This section presents experiments to show the capabilities of our framework and to support the four key contributions discussed at the beginning of the paper and repeated below for the reader's convenience: 
\begin{enumerate}[start=1,label={\bfseries C\arabic*}]
    \item A synchronization protocol that ensures consistent simulation behavior even beyond real-time speeds.
    \item A modular design that \rev{can }support various robotic systems and state, action, and time-scale abstractions.
    \item An agnostic design that allows compatibility with multiple engines.
    \item \rev{The }integrated delay simulation and domain randomization \rev{features } in EAGERx can narrow the sim2real gap.
\end{enumerate}

\subsection{Experimental Setup}
\begin{figure*}[t]
    \centering
    \begin{subfigure}[b]{0.175\linewidth}
        \centering
        \includegraphics[width=0.91\textwidth]{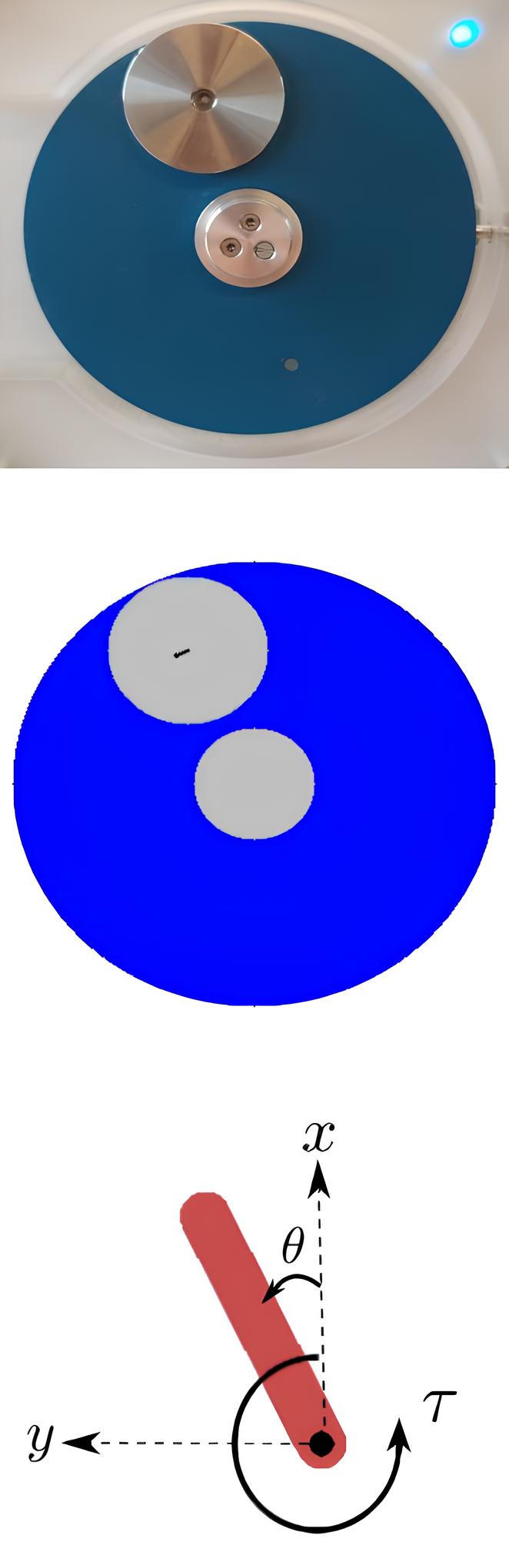}
        \caption{Swing up}
        \label{fig:setup_pendulum}
    \end{subfigure}
    \begin{subfigure}[b]{0.25\linewidth}
        \centering
        \includegraphics[width=0.91\textwidth]{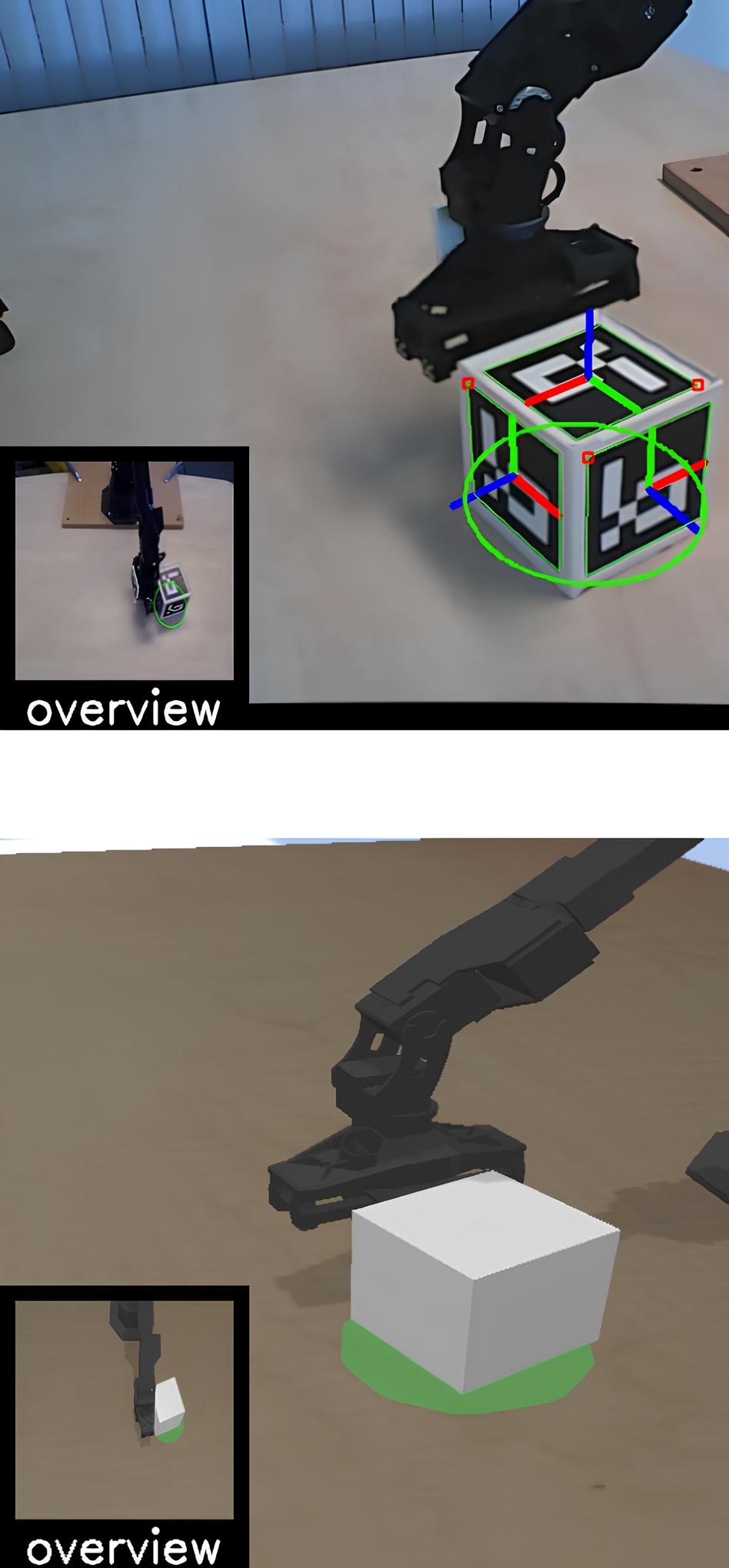}
        \caption{Box pushing}
        \label{fig:setup_box}
    \end{subfigure}
        \begin{subfigure}[b]{0.56\linewidth}
        \centering
        \includegraphics[width=0.91\textwidth]{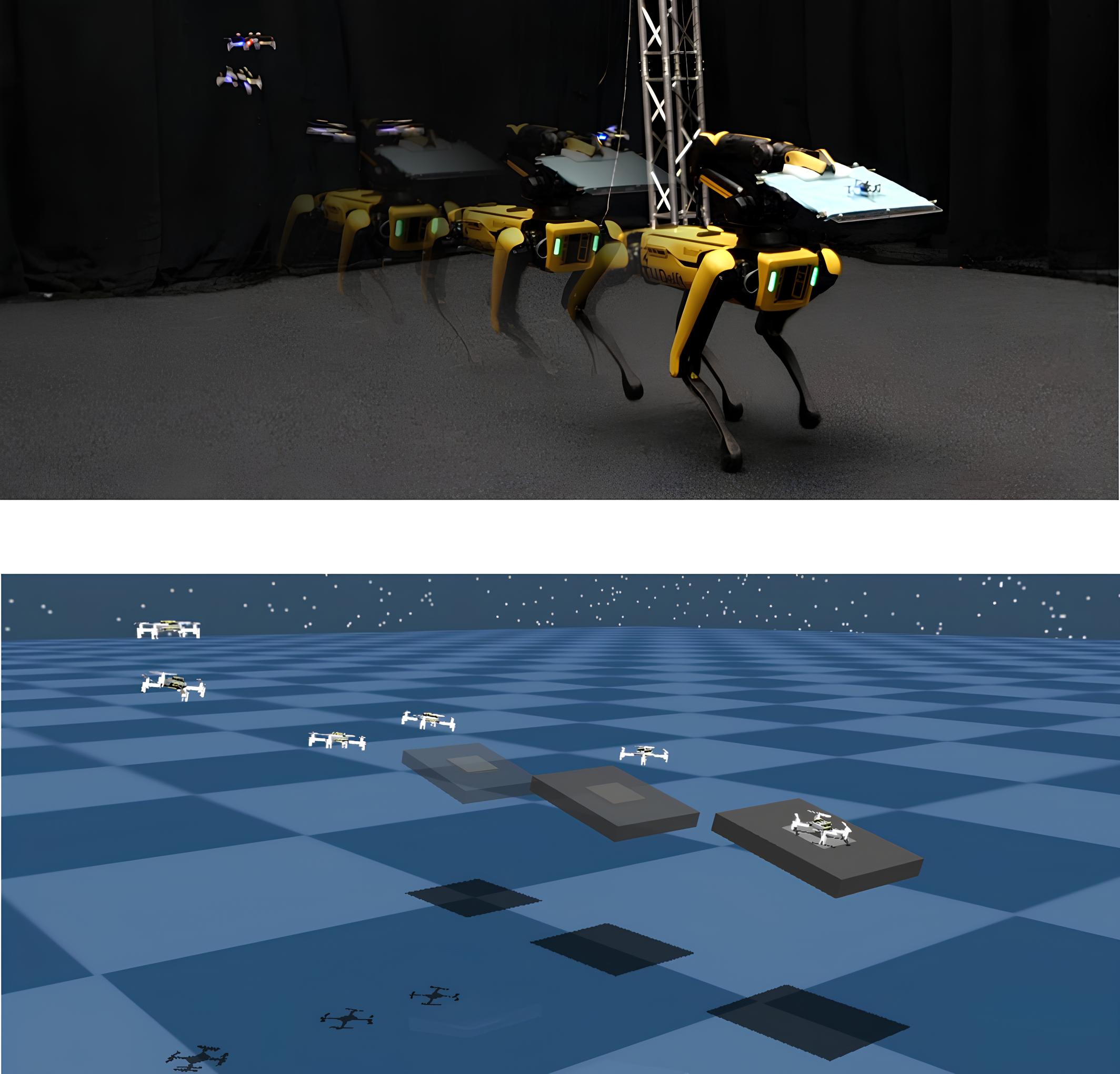}
        \caption{Inclined landing}
        \label{fig:setup_cf}
    \end{subfigure}
    \caption{\rev{
    Diverse robotic system tasks illustrating the EAGERx framework's flexibility. 
    (a) Swing-up task with an inverted pendulum, highlighting delay compensation in reinforcement learning. The task involves zero-shot evaluations on a real-world pendulum setup, comparing a disk-based simulator with the OpenAI Gym rod-based environment. 
    (b) Box-pushing experiment using a Viper 300x robotic manipulator, emphasizing the need for domain randomization with a low-resolution Logitech C170 webcam for box localization tracking. 
    (c) Inclined landing task where a quadrotor lands on a moving and inclined deck, showcasing the integration of multiple mobile robots into a dynamic task.
    }}
    \vspace{\vspacebelow}
    \label{fig:setups}
\end{figure*}

\rev{EAGERx is validated with a pendulum swing-up task, a vision-based box-pushing task, and an inclined landing experiment for a quadrotor.
The simulated and real-world setups of all three tasks are depicted in \figref{fig:setups}. 
To validate \textbf{C1}, we experimentally assess \algref{alg:protocol} and employ it in accelerated, parallelized training for all tasks.
Claims \textbf{C2-C3} are validated by the tasks involving different \textit{engines} and distinct types of systems like pendulums, manipulators, quadrotors, and quadrupeds.
Claim \textbf{C4} is validated by demonstrating the detrimental effect of delays and model mismatch on sim2real performance and showing how simulating delays and domain randomization can restore sim2real performance.
All policies are trained in simulation, and zero-shot evaluated on their real-world counterparts.
}

\textbf{Swing Up}
The inverted pendulum task addresses the classic control problem of swinging up and stabilizing an underactuated pendulum. 
\rev{The choice of this task is intentional; it emphasizes the critical challenge of delay compensation in reinforcement learning.
By showing how ignoring delay simulation can hinder policy transfer even in straightforward scenarios, we highlight the significant consequences for more complex systems where delays are inevitable and complexity is higher.
The simplicity of the task underscores the fundamental importance of addressing delays in sim2real approaches. }
We conduct zero-shot evaluations using a real-world pendulum setup comprising a mass on a disk driven by a DC motor.
To train policies, we utilize two simulators.
The first simulator's dynamics model aligns with the physical system, representing the pendulum as a disk.  
In contrast, the second simulator adopts the OpenAI Gym Classic Control's \textit{Pendulum} environment, modeling the pendulum as a rod \citep{brockman2016openai}, inadvertently introducing a sim2real gap that requires mitigation. 
\rev{All three systems are depicted in \figref{fig:setup_pendulum}. }
In all experiments, the pendulum is controlled at a rate of 20\,Hz, while sensor measurements are obtained at a rate of 60\,Hz.
\rev{
    The agent observes the last two received sensor measurements.
    Users can specify such a rolling window length when connecting nodes in EAGERx, as shown in lines 14 and 16 of \coderef{code:main}).
}
\rev{Policies are trained using the soft actor-critic (SAC) \cite{haarnoja2019soft} implementation from \cite{raffin2021stable}.}

\textbf{Box Pushing}
\rev{In the box-pushing experiment, a Viper~300x robotic manipulator moves a box to a target based on streaming webcam images.
To emphasize the importance of domain randomization, we use a consumer-grade Logitech~C170 webcam, selected for its low resolution, modest frame rate, and high latency, to track the box's position and orientation. }
For evaluation, we selected six unique initial configurations (three positions approximately $30$\,cm from the goal for both a yaw angle of 0 and $\frac{\pi}{2}$\,rad) and repeated them thrice per policy.
\rev{Policies are trained in PyBullet using the SAC \cite{haarnoja2019soft} implementation from \cite{raffin2021stable} with hindsight experience replay \cite{andrychowicz2017hindsight}. 
The simulation and real-world setups are shown in \figref{fig:setup_pendulum}. 
}

\rev{\textbf{Inclined Landing}
To demonstrate the framework's ability to facilitate control in highly dynamic environments, we trained an agent to perform the challenging maneuver of landing a quadrotor on an inclined and moving landing deck. 
Due to the configuration of its rotors, standard quadrotors can only exert thrust upwards, as rotor spinning directions cannot be reversed mid-flight. 
This under-actuation complicates landing on an incline, as the agent can only decelerate when approaching the deck. 
Therefore, if the agent initiates the landing procedure with insufficient momentum, it cannot accelerate, resulting in a crash.
In \cite{kooi2021inclined}, PPO \cite{schulman2017proximal} was used to learn a policy for landing on a stationary landing deck in 2D ($xz$-plane) with a fixed inclination ($25^\circ$). 
In this paper, we follow a similar approach using the PPO implementation from CleanRL \cite{huang2022cleanrl}. 
However, we extend the policy's capability to land in 3D ($xyz$-plane), at various inclinations ($0-25^\circ$), and on a moving landing deck ($0-1 \text{m/s}$). 

As in \cite{kooi2021inclined}, the quadrotor dynamics are prescribed by ODEs identified with real-world data. 
In simulation, the landing deck moves in a straight line at a fixed inclination, varying speed, inclination, and direction across episodes to learn multi-goal behavior. 
To model the interaction between the landing deck and quadrotor, we extend the ODE dynamics with MuJoCo's \cite{todorov2012mujoco} collision detection capabilities to detect successful landings and crashes.
During real-world evaluation, we move the landing deck around with a quadruped and track the pose of both the deck and quadrotor with an accurate motion capture system.
The simulation and real-world setups are shown in \figref{fig:setup_cf}. 
}

\subsection{Analysis}
\begin{figure}[t]
    \centering
    \includegraphics[width=\linewidth]{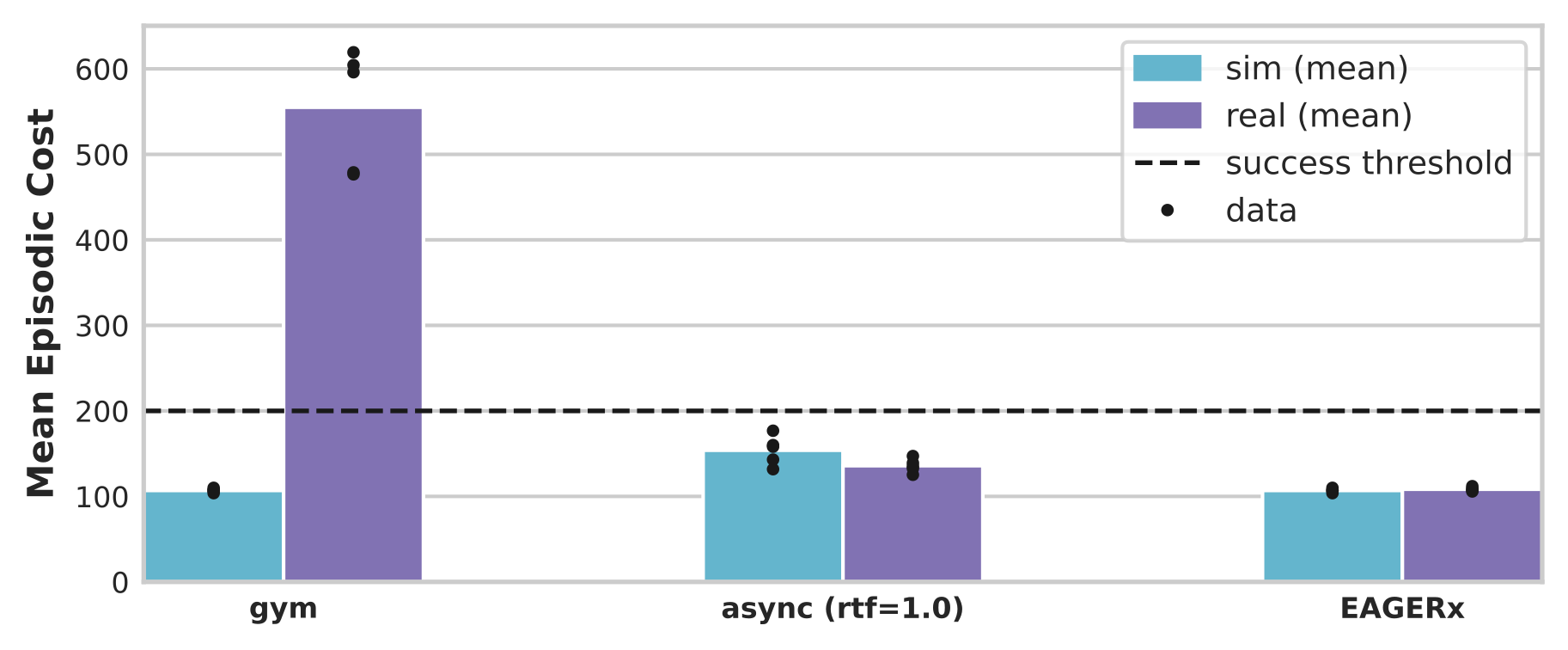}
    \caption{
    Comparison of mean episodic cost between simulations and real-world pendulum performance. 
    The success threshold denotes the level below which a 100\% success rate is achieved. 
    Performance drops notably in the real-world scenario with a conventional \emph{gym} approach, illustrating the sim2real gap.
    Asynchronous simulation (\emph{async}) at real-time speeds mitigates the gap but leads to excessively long training times.
    Synchronized training under our protocol (\emph{EAGERx}) facilitates consistent performance at faster-than-real-time simulation speeds.
}
    \label{fig:sync_vs_async}
    \vspace{\vspacebelow}
\end{figure}

\begin{figure}[t]
    \centering
    \centering
    \includegraphics[width=\linewidth]{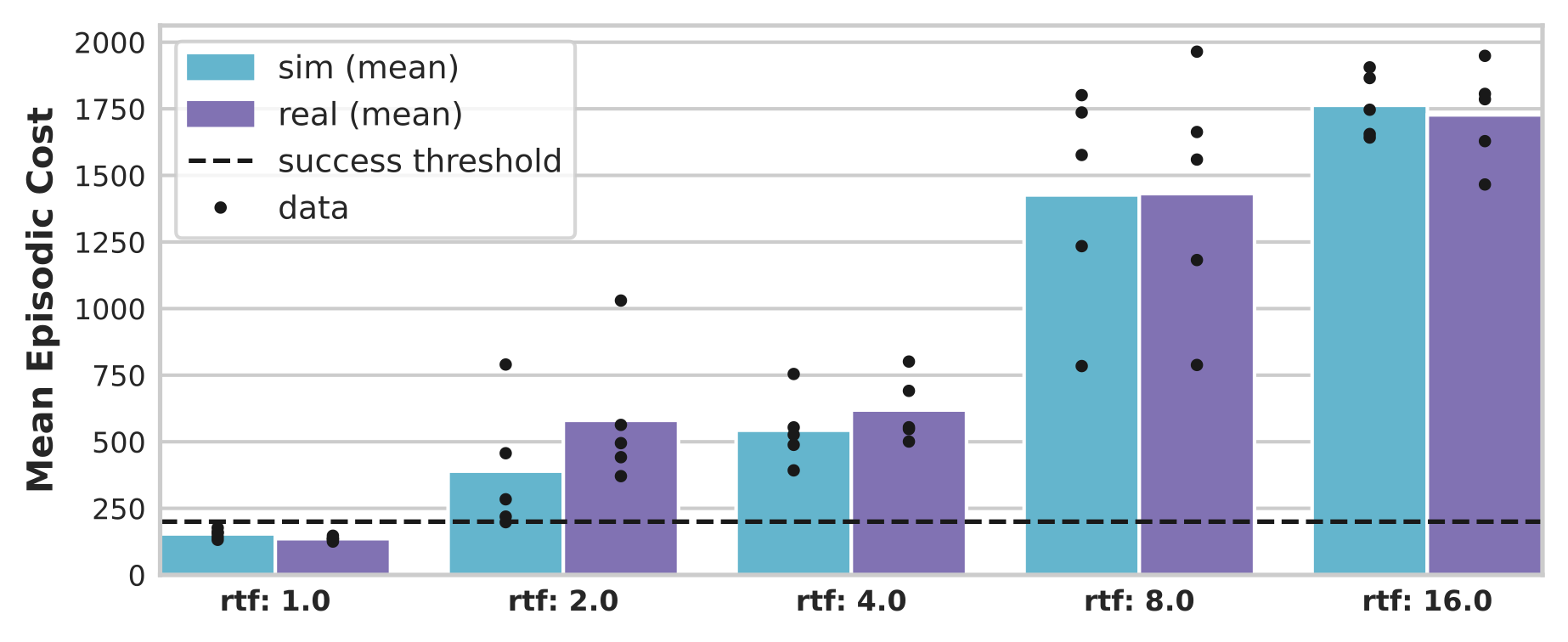}
    \caption{
        The impact of varying real-time factors (rtf) on the mean episodic cost in a simulated pendulum environment. 
        Performance declines as the rtf increases, indicating the challenges of maintaining fidelity in faster-than-real-time simulations when components operate asynchronously.
    }
    \label{fig:rtf_pendulum}
    \vspace{\vspacebelow}
\end{figure}

\begin{figure}[t]
    \centering
    \begin{subfigure}[b]{0.49\linewidth}
        \centering
        \includegraphics[width=\textwidth]{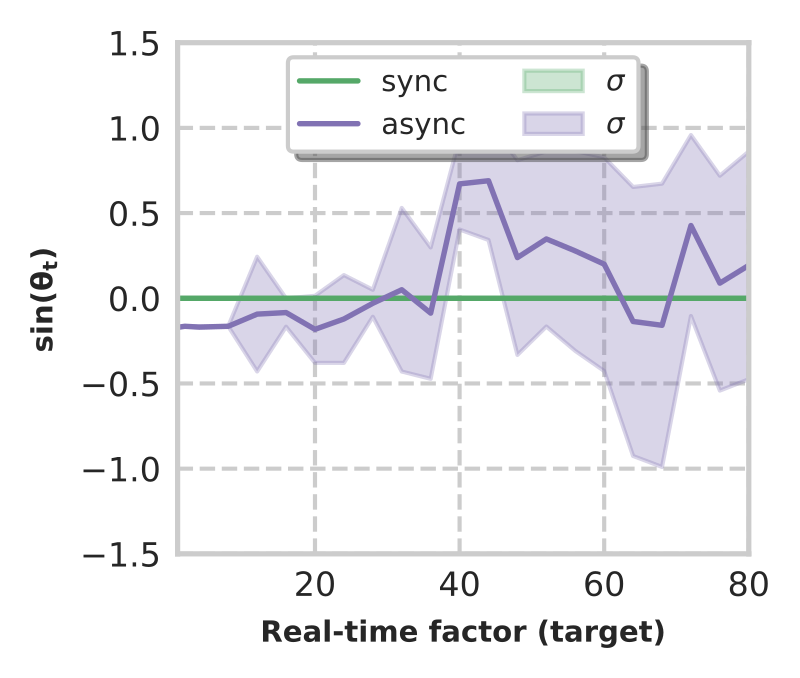}
        \caption{Variation in angle $\sin(\theta)$}
        \label{fig:error}
    \end{subfigure}
    \begin{subfigure}[b]{0.46\linewidth}
        \centering
        \includegraphics[width=\textwidth]{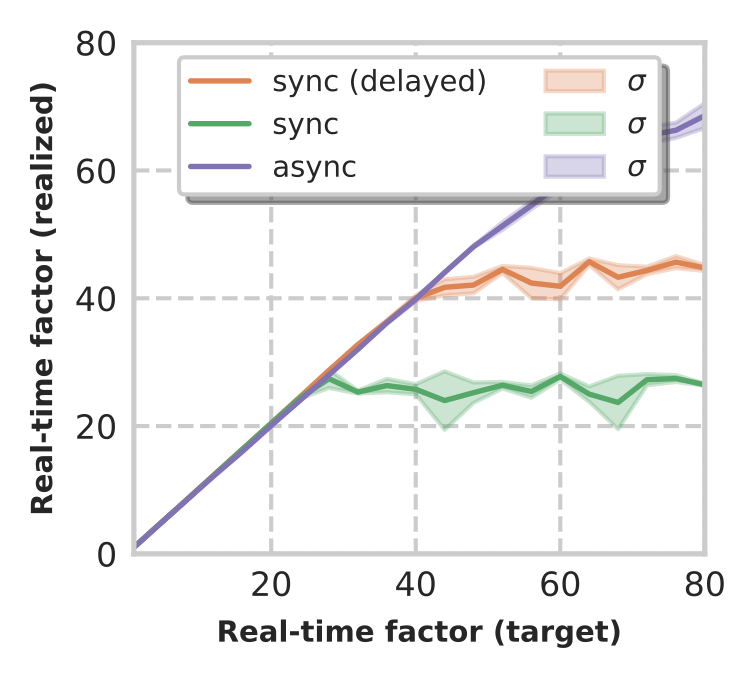}
        \caption{Real-time factor}
        \label{fig:rtf}
    \end{subfigure}
    \caption{A comparison between asynchronous (\textit{async}) and synchronous (\textit{sync}) simulations of a pendulum at faster-than-real-time speeds.
    \figref{fig:error} shows the variation in angle $\sin(\theta)$ at $t=2.0$\,s over 5 runs of a simulated pendulum as a function of the real-time factor.
    \figref{fig:rtf} shows the realized real-time factor of the simulation for both synchronous and asynchronous cases.
    }
    \vspace{\vspacebelow}
    \label{fig:protocol}
\end{figure}

\textbf{C1} We tested \algref{alg:protocol}'s ability to maintain consistent simulation behaviors at speeds surpassing real-time, using experiments with the disk pendulum. 
Initially, we utilized the disk simulator within a standard OpenAI Gym environment, trained a policy, and then conducted a zero-shot evaluation on the actual system. 
As depicted in \figref{fig:sync_vs_async}, the performance significantly declines in the real world, indicating a substantial simulation-to-reality (sim2real) gap. 
This discrepancy results from the sequential communication in simulations contrasted with the asynchronous sensor and actuator commands in the real system via ROS topics \citep{quigley2009ros}, forcing the agent to sometimes rely on outdated information in real-world scenarios. 
To mimic this asynchronous nature, we adapted the gym environment to use asynchronous communication in simulation. 
This adaptation enabled the policy to handle occasional delays, enhancing its real-world applicability. 
However, this required limiting simulation speed to a real-time factor of 1, considerably prolonging training duration. 
The real-time factor, the ratio of simulation to real-world time, at 1 signifies running the simulation in real-time. 
\figref{fig:rtf_pendulum} demonstrates that increasing this factor degrades performance, underscoring the challenges in accelerating simulation beyond real-time while ensuring effective real-world transfer.

Setting an excessively high target real-time factor may cause parallel components in the simulation to desynchronize and lag. 
Figures \ref{fig:error} and \ref{fig:rtf} demonstrate the consequences of this lag in asynchronous simulations. 
Specifically, \figref{fig:error} displays the variation in the simulated pendulum's angle, $\sin(\theta)$, at $t=2.0$\,s across five runs with identical input sequences, highlighting the increasing discrepancy in angle measurements as the real-time factor rises. 
In contrast, simulations synchronized via our protocol remain deterministic while still allowing parallel operations, enhancing speed without sacrificing accuracy. 
Synchronous simulations naturally cap the real-time factor to preserve synchronization, whereas asynchronous ones might show a misleadingly high factor, as evidenced in \figref{fig:error}, where increased speed incurs greater variability and component desynchronization. 
This illustrates the adverse effects of unsynchronized, accelerated simulations. 
Adapting the disk simulator into an EAGERx environment for synchronized training under our protocol facilitated faster-than-real-time speeds while ensuring consistency between simulated and real-world behaviors, as depicted in \figref{fig:sync_vs_async}.

\begin{figure*}[t]
    \centering
    \begin{subfigure}[b]{0.43\linewidth}
        \raggedright
        \includegraphics[width=0.99\textwidth]{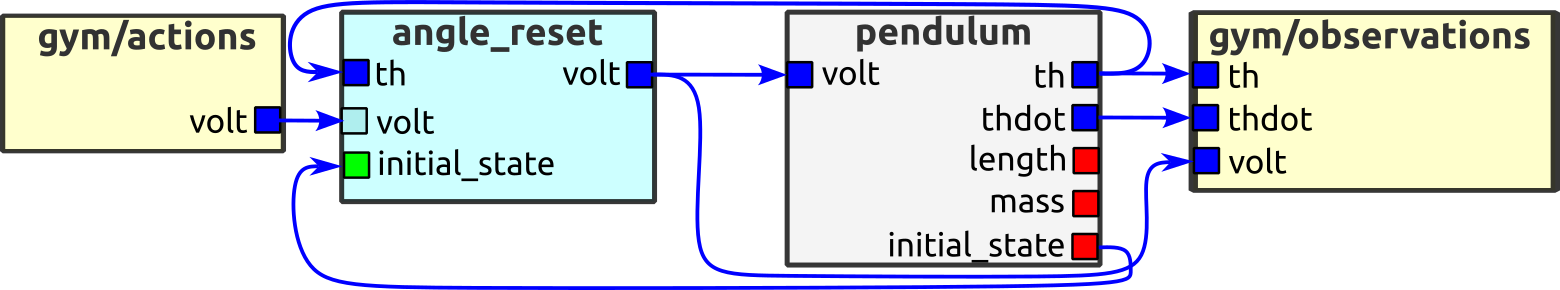}
        \caption{Agnostic graph - swing up}
        \label{fig:pendulum_graph}
    \end{subfigure}    
    \begin{subfigure}[b]{0.56\linewidth}
        \raggedleft
        \includegraphics[width=0.99\textwidth]{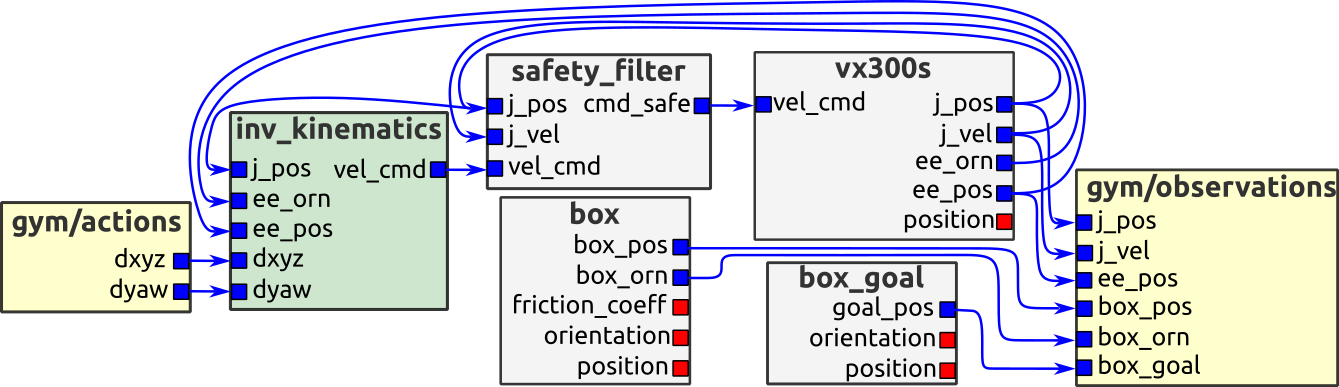}
        \caption{Agnostic graph - box pushing}
        \label{fig:box_push_graph}
    \end{subfigure}%
    \hfill
    \par\smallskip \par\smallskip
    \begin{subfigure}[b]{0.43\linewidth}
        \raggedright
        \includegraphics[width=0.99\textwidth]{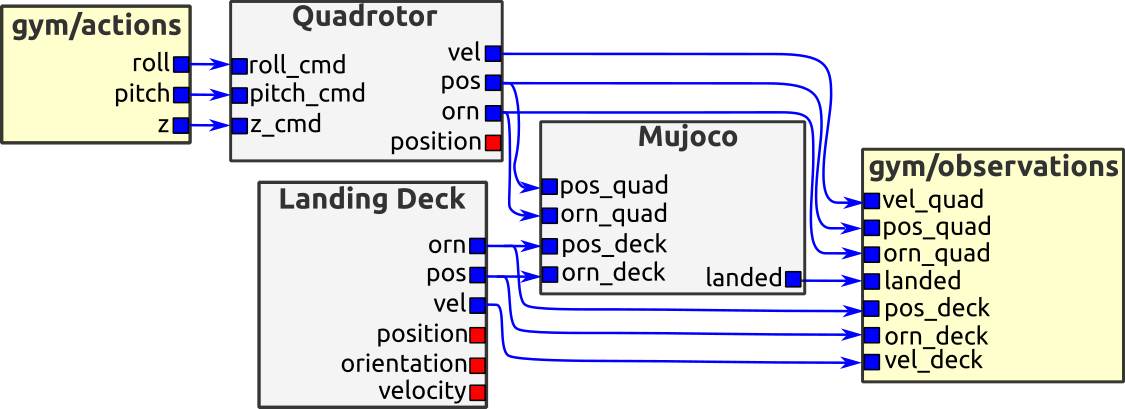}
        \caption{Agnostic graph - inclined landing}
        \label{fig:cf_graph}
    \end{subfigure}
    \begin{subfigure}[b]{0.275\linewidth}
        \centering
        \includegraphics[width=0.90\textwidth]{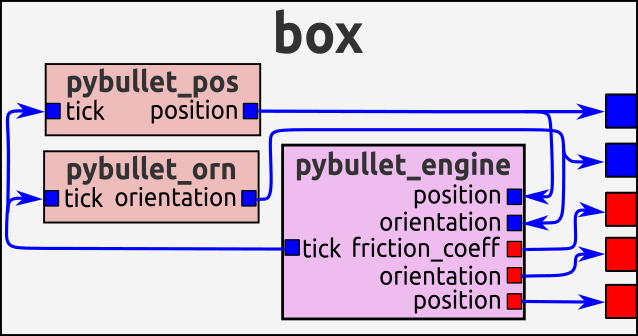}
        \caption{PyBullet subgraph}
        \label{fig:box_graph_pybullet}
    \end{subfigure}
    \begin{subfigure}[b]{0.274\linewidth}
        \centering
        \includegraphics[width=0.90\textwidth]{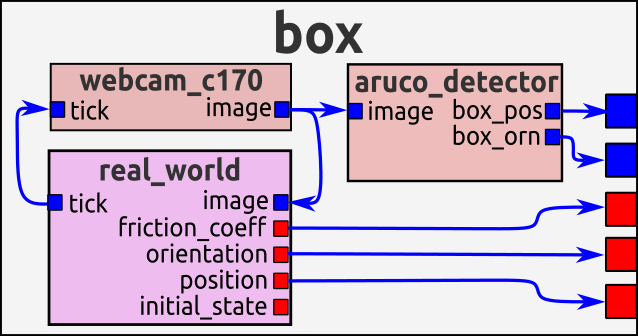}
        \caption{Real world subgraph}
        \label{fig:box_graph_real}
    \end{subfigure}
    \caption{\rev{
    Diverse robotic system tasks demonstrating the versatility of EAGERx's graph-based design. 
    (a) The pendulum swing-up task uses an agnostic graph with an \textit{angle reset} node for initializing the pendulum's position. 
    (b) The vision-based box-pushing task utilizes an \textit{inverse kinematics} node for task-space learning and a \textit{safety filter} for correcting hazardous commands. 
    The engine-specific subgraphs for replacing the \textit{box} object in (b) are depicted for the PyBullet (d) and real-world (e) engines.    
    (c) The inclined landing task illustrates how EAGERx integrates collision detection in MuJoCo with ODE dynamics to get the best of both simulators.
    }}
    \vspace{\vspacebelow}
    \label{fig:graphs}
\end{figure*}

Our protocol, designed for robotic system synchronization, does not necessitate synchronous operation within the simulation. 
In fact, asynchronous communication still permits nodes to transmit messages and perform tasks without waiting for immediate responses, thereby accelerating the simulation and allowing nodes to progress based on their processing capabilities and data availability. 
This is illustrated in \ref{fig:rtf}, where we introduced a simulated delay between the pendulum actuator and the physics engine.
Consequently, the pendulum's callback and the physics engine's callback can be executed concurrently, as the physics engine's callback relies on the pendulum's output from the previous timestep rather than the current one. 
Since each node's protocol operates independently, this parallelization occurs naturally, resulting in approximately $50\%$ increase in the realized real-time factor for the synchronized simulation compared to the case without delay.

\rev{\textbf{C2} To support the claimed contribution that the framework accommodates various robotic systems, the tasks involve distinct robot systems such as pendulums, manipulators, quadrotors, and quadrupeds. }
EAGERx's graph-based design, enabling diverse abstractions, is demonstrated in the vision-based box-pushing task.
Rather than end-to-end training on raw images, an \textit{aruco detector} is used for state abstraction as depicted in \figref{fig:box_graph_real}, negating the need for photorealistic rendering.
Action abstractions, visible in \figref{fig:box_push_graph}, include an \textit{inverse kinematics} node for task-space learning and a \textit{safety filter} correcting hazardous commands. 
Nodes set at optimal rates ensure efficient resource use and learning. 
The pendulum task underlines the framework's modularity using an \textit{angle reset} node, visible in \figref{fig:pendulum_graph}, to position the pendulum at the initial angle via PID control before a new episode.
\rev{Finally, we demonstrate EAGERx's capability to coordinate diverse systems, such as a quadruped and quadrotor, in a delay-sensitive and dynamic task with the inclined landing experiment. }

\textbf{C3} To support the claimed contribution that EAGERx is compatible with a variety of physics engines and the real world, we conducted experiments with four different \textit{engines}---PyBullet \cite{coumans2021pybullet}, OpenAI Gym Classic Control \cite{brockman2016openai}, real-world, and simulations with sets of ODEs---showing the ability to switch between real and simulated counterparts. 
The box-pushing task demonstrates how a division of the graph into engine-specific and engine-agnostic subgraphs resulted in a unified pipeline between PyBullet and reality.
The \textit{inverse kinematics} and \textit{safety filter} nodes work with any simulator, as seen in the agnostic graph (\figref{fig:box_push_graph}), while the \textit{aruco detector} and \textit{webcam} nodes are swapped with PyBullet-specific nodes in Figures \ref{fig:box_graph_pybullet} and \ref{fig:box_graph_real}.
Likewise, the agnostic graph in \figref{fig:pendulum_graph} was used in all pendulum experiments to display sim2real transfer across physics engines.
\rev{The inclined landing experiment further illustrates the framework's flexibility by combining the collision detection capabilities of MuJoCo \cite{todorov2012mujoco} with the accurately identified ODE dynamics of the quadrotor. 
This task highlights how different physics engines can be integrated seamlessly within EAGERx. 
The collision detection in MuJoCo is used to detect successful landings and crashes, while the ODE dynamics ensure realistic quadrotor behavior.
Collision detection is used both in simulation and reality, so it is therefore placed in the agnostic graph \figref{fig:cf_graph}.
During real-world evaluation, the landing deck is moved by a quadruped and the poses of both the deck and the quadrotor are tracked using a motion capture system. 
Since simulating the full dynamics of a quadruped during policy learning is unnecessary and would only slow down training with redundant computation, the quadruped control nodes are placed in the real-world engine-specific graph. 
We can simulate just a moving landing platform without the quadruped, as actuation is not required to move objects in simulation. 
This approach focuses computational resources on what truly matters for training. 
}

\begin{figure}[t]
    \centering
    \includegraphics[width=\columnwidth]{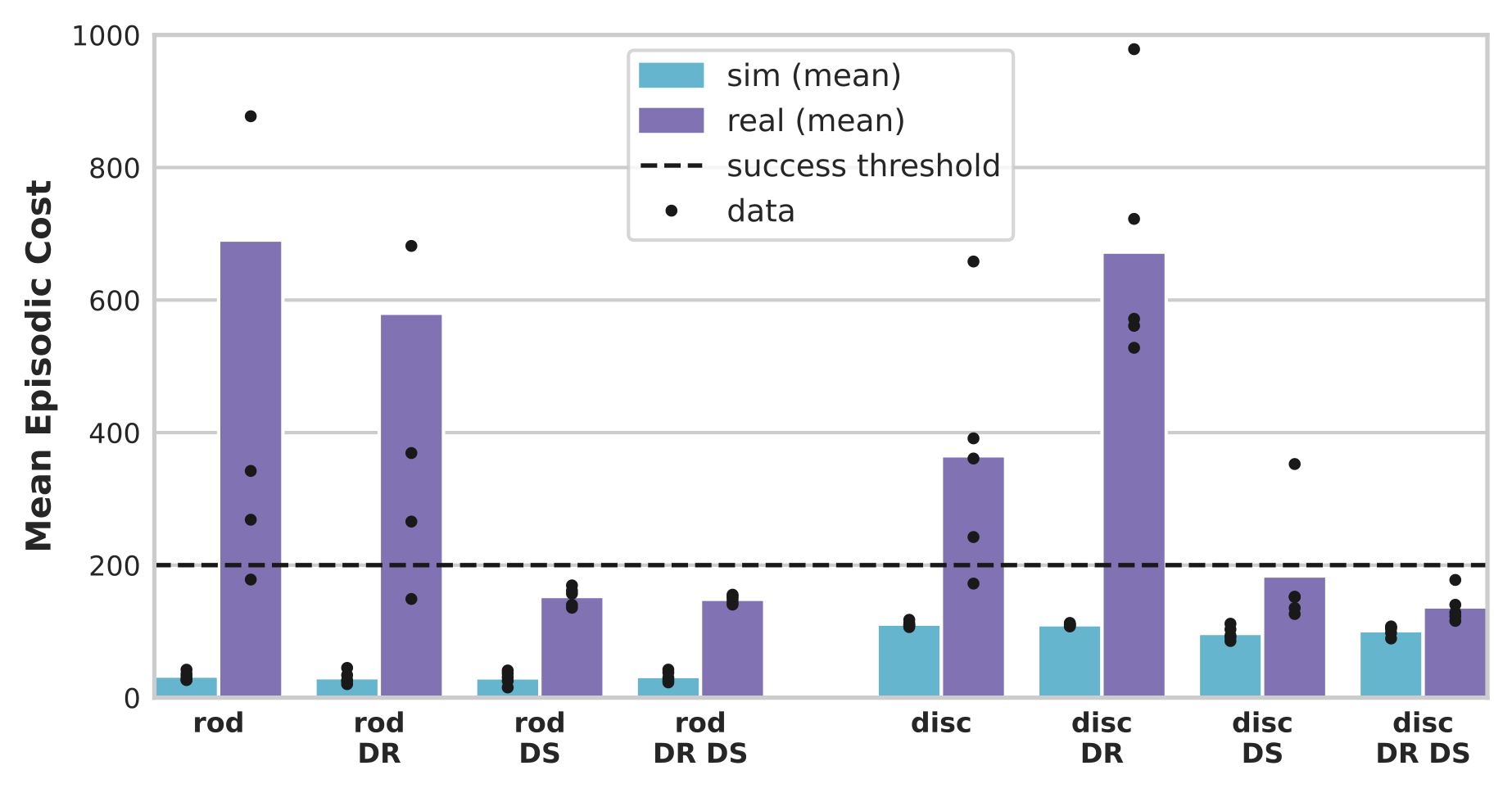}    
    \caption{
        \rev{Results for the pendulum swing-up task show the mean episodic cost for 5 policies (10 episodes per policy) and the impacts of domain randomization (DR) and delay simulation (DS). 
        Here \textit{rod} and \textit{disk} refer to the engine used during training, as depicted in \figref{fig:setup_pendulum}. }
        For clarity, the y-axis is capped at 1000, though note this truncates some data points. 
        The \textit{success threshold} indicates 100\% success rate, meaning successful pendulum swing up and stabilization each episode for all evaluations below this threshold. 
    }
    \label{fig:pendulum_results}
    \vspace{\vspacebelow}
\end{figure}

\begin{figure}[t]
    \centering
    \includegraphics[width=\columnwidth]{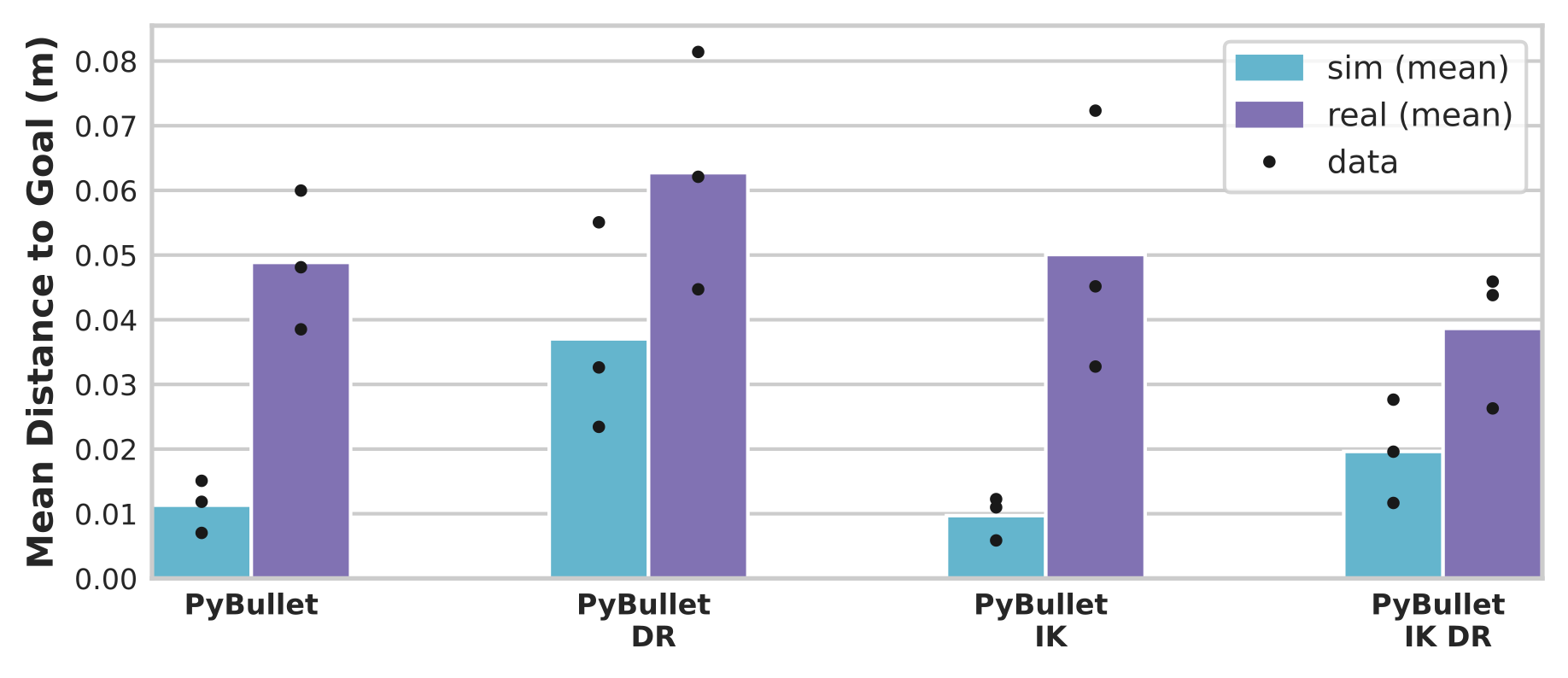}
    \caption{    
    \rev{Results for the box-pushing task }show the mean distance from the goal at the end of 16 episodes for 3 policies and evaluate the benefits of an inverse kinematics (IK) node (facilitating task space control) and domain randomization (DR) of the friction coefficient. 
    }
    \vspace{\vspacebelow}
    \label{fig:box_push_results}
\end{figure}

\begin{figure}[t]
    \centering
    \includegraphics[width=\columnwidth]{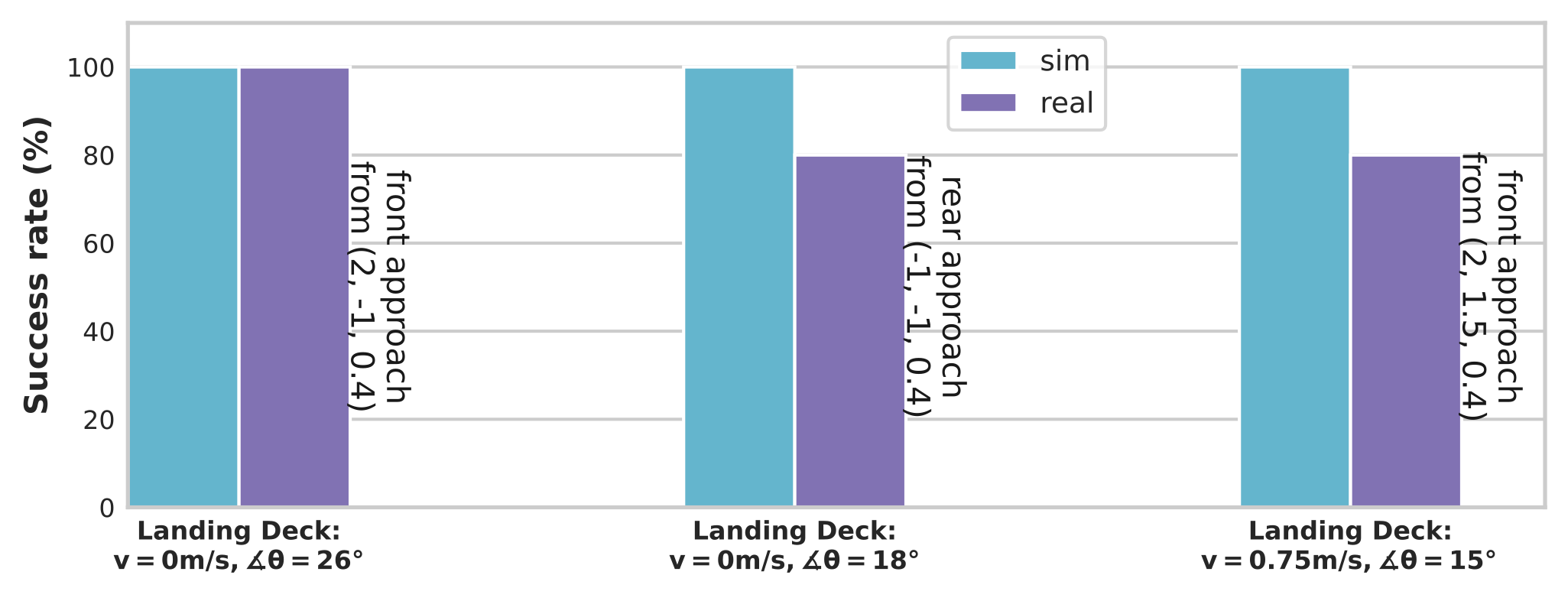}
    
    \caption{\rev{Results for the inclined landing experiment show the success rate for landing on a stationary and moving deck at various inclinations in simulation and real-world settings. 
    The experiments evaluate the performance of the policy in terms of successful landings across 10 episodes, demonstrating the framework's capability to handle dynamic and delay-sensitive tasks involving diverse robotic systems like quadrupeds and quadrotors. }}
    \vspace{\vspacebelow}
    \label{fig:cf_results}
\end{figure}

\textbf{C4} We show that the integrated delay simulation and domain randomization features can reduce the sim2real gap by demonstrating that the negative impacts of actuator delay can be counteracted using the delay simulation feature during training for two different simulated versions of the pendulum.
In this task, we supported \textbf{C4} by evaluating policies on the real system with an actuator delay set at the smallest value that led to a breakdown in baseline performance.
When we progressively increased the actuator delay, it resulted in baseline policy failure for delays of $0.025$\,s and $0.035$\,s for the rod and disk pendulum, respectively.
Our experiments studied the potential of training with domain randomization and/or delay simulation to mitigate the adverse effects of the actuator delay.
For the disk pendulum, we applied randomization within $\pm 10\%$ of the mean values ($0.033$\,kg for mass and $0.1$\,m for length).
For the rod pendulum, randomization was limited to $\pm 5\%$, considering the higher accuracy of this model.
Delay simulation involved randomization within $\pm 0.005$\,s around the set actuator delay.
The results shown in \figref{fig:pendulum_results} suggest that delay simulation can mitigate the adverse effects of actuator delay for zero-shot transfer from both the rod and disk simulator to the real pendulum system.
In the disk scenario, adding domain randomization to delay simulation further improved performance and resulted in successful transfer with the smallest performance gap between simulation and reality.
The effectiveness of domain randomization is further highlighted in the box-pushing task (\figref{fig:box_push_results}). 
We examined its impact by altering the box's friction coefficient between 0.1 and 0.4. 
\figref{fig:box_push_results} shows that, compared to the baseline, friction randomization reduces the performance gap between simulation and reality despite lowering overall performance, thereby illustrating that relying solely on domain randomization can increase task difficulty.
Conversely, incorporating the inverse kinematics node combined with friction randomization enhances performance while reducing the gap between simulation and real-world execution.
\rev{
We used a delay simulation of 0.02 seconds for the inclined landing task and randomized the mass within $\pm 5\%$, leading to the results in \figref{fig:cf_results}. 
However, we refrain from conducting an extensive ablation study on the effects of delay simulation and domain randomization to avoid unnecessary hardware damage.
}

\section{Applications beyond Reinforcement Learning}
\label{sec:beyond_rl}
The modular design and unified software pipeline of the framework have utility in various other domains.
This section explores two such instances: interactive imitation learning and Machine Learning (ML) enhanced classical control, showcasing EAGERx's utility beyond reinforcement learning.

\subsection{Interactive Imitation Learning}
\begin{figure*}[t]
    \centering
    \includegraphics[width=\linewidth]{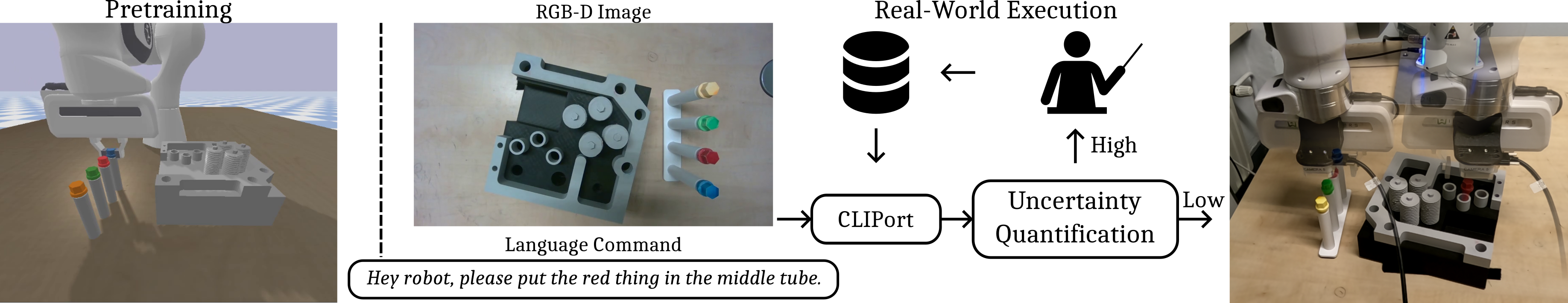}
    \caption{In this application of EAGERx, a CLIPort \citep{shridhar2022cliport} model is trained using an active learning approach that queries the human teacher for a demonstration in case of high prediction uncertainty.
    Also, the model is pre-trained using demonstrations gathered in simulation.}
    \label{fig:diesel_engine}
\end{figure*}

This application shows how EAGERx is suitable for (interactive) imitation learning.
Here, the task involves assembling a mock-up Diesel engine by following voice commands from a human operator.
The parts used in this task are 3D-printed versions of the parts from an actual Diesel engine assembly setup. 
To solve this task, we apply a learning from demonstration approach based on CLIPort \citep{shridhar2022cliport}.
However, we utilize an interactive imitation learning approach instead of gathering offline demonstrations only.
Collecting on-policy data helps us to, for example, learn to recover from failures.
Learning recovery behavior is often not possible using demonstrations collected offline by experts since they are unlikely \rev{to } visit failure states.
We apply an active learning method based on uncertainty quantification \citep{luijkx2022partnr}.
This method actively queries the human teacher for a demonstration in case there is high prediction uncertainty.
EAGERx offers three main advantages in this scenario.
First of all, we can easily create a digital twin of the real-world environment in simulation.
This allows one to debug a large portion of the pipeline in simulation, which is safe and time-efficient.
Moreover, the simulated environment facilitates the cost-effective collection of synthetic demonstrations.
These can be used to pre-train the policy in simulation and speed up learning.
Lastly, EAGERx's modular graph structure enables the simple connection of various components.
In this case, the graph includes a speech-to-text transcriber, the policy node, as well as the RGB-D camera, and the manipulator.
An overview of the task is shown in \figref{fig:diesel_engine}.
A video demonstration of this application is available at \url{https://eagerx.readthedocs.io}.

\subsection{ML-Enhanced Classical Control}
\begin{figure}[t]
    \centering
        \begin{subfigure}[b]{0.49\linewidth}
        \centering
        \includegraphics[width=\textwidth]{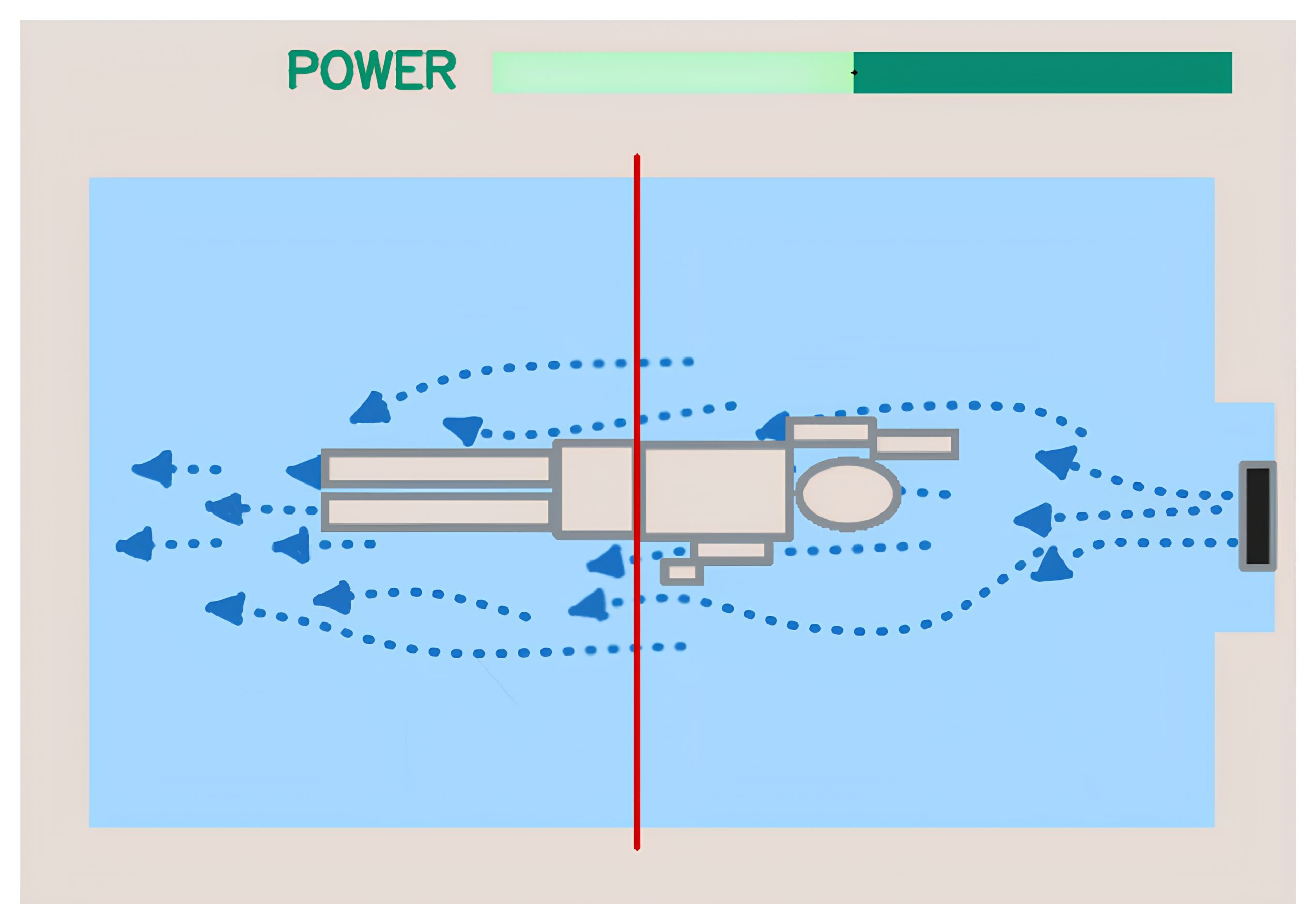}
        \caption{Custom simulator}
        \label{fig:swim_sim}
    \end{subfigure}
    \begin{subfigure}[b]{0.49\linewidth}
        \centering
        \includegraphics[width=\textwidth]{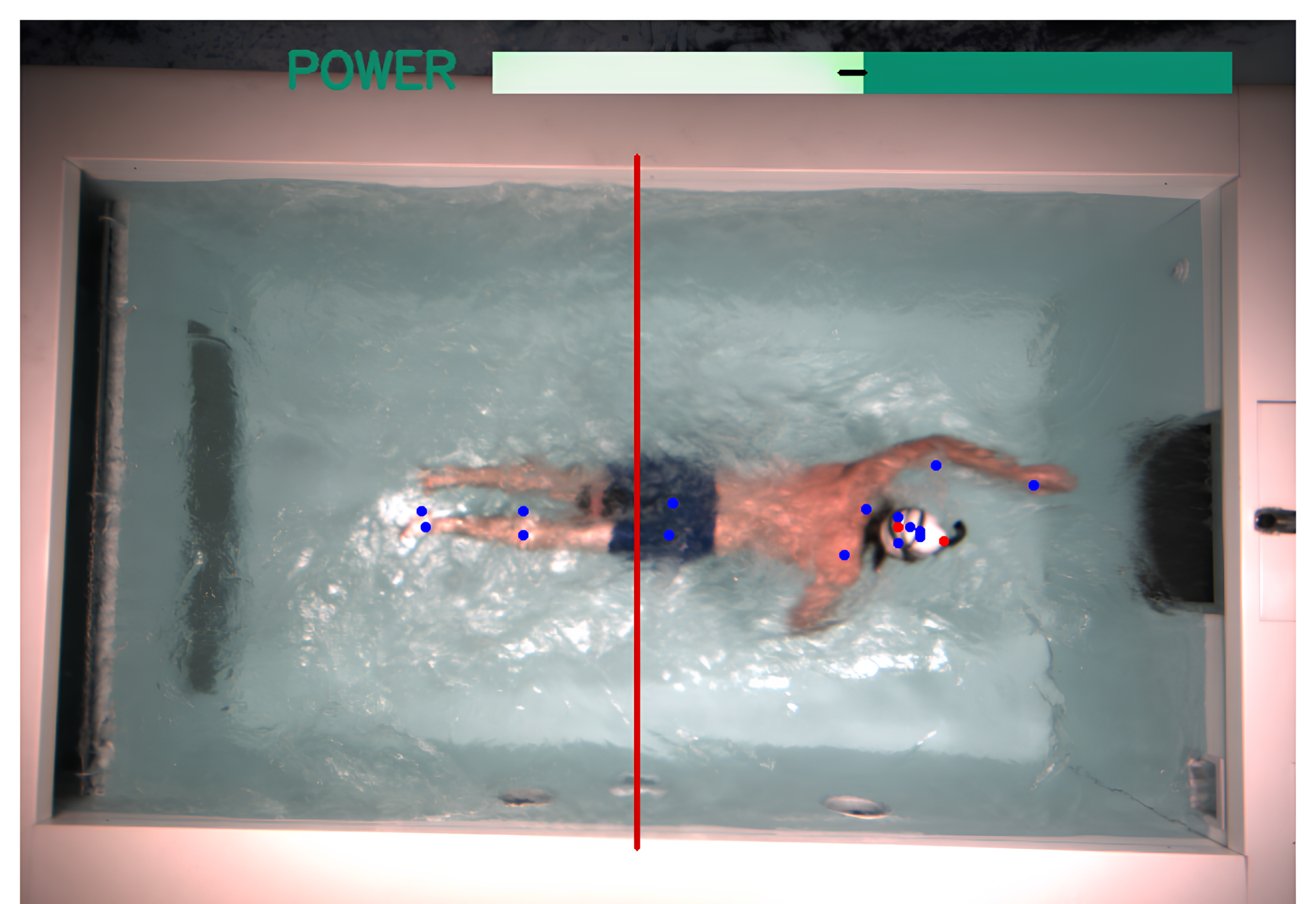}
        \caption{Real}
        \label{fig:swim_real}
    \end{subfigure}
    \caption{Application of EAGERx in an adaptive swimming pool environment. 
    The system modulates the pool's counter-current in response to the swimmer's hip position relative to a preset boundary (red line), utilizing a pose detector and Kalman filter for position estimation and a PID controller for current adjustment.}
    \vspace{\vspacebelow}
    \label{fig:swim}
\end{figure}

This application illustrates EAGERx’s integration of pre-trained ML models with classical control in a custom simulator, addressing a practical challenge. 
EAGERx was applied to an adaptive swimming pool environment, showcased in \figref{fig:swim}. 
This environment enhances traditional counter-current pools by dynamically adjusting the current based on the swimmer's position. 
Normally, it is the swimmer’s task to stay centered in the pool, a difficult task for beginners. 
Our approach, however, modifies the pool's counter-current in line with the swimmer's location, maintaining central positioning regardless of swim speed. 
This adaptability makes the pool more user-friendly for novice swimmers.

Variable transport delays complicate the control problems. 
Specifically, alterations in motor power do not instantaneously translate into flow changes; this delay results from the gradual response of the water pump's first-order dynamics, as well as the variable time it takes for a change in water flow to impact the swimmer, contingent on their position in the pool. 
When the swimmer is towards the front, they feel the effects of flow velocity changes more rapidly than when positioned at the rear. 
The absence of a readily available off-the-shelf simulator for this specific scenario underscored the utility of EAGERx, which facilitated the creation of a custom simulator, proving invaluable in the development of the control pipeline.

The modular architecture of EAGERx facilitated the integration of a pose detector with a Kalman filter, resulting in estimates of the swimmer's position and velocity from solely top-view camera imagery. 
Subsequently, a PID controller was employed to modulate the pool current in alignment with these estimates. 
A video demonstration is available in the supplemental material.

%% file: discussion.tex
\section{Discussion}
\label{sec:discussion}

Comparing EAGERx with ROS \citep{quigley2009ros} might seem natural due to their modular structures and asynchronous communication.
Nonetheless, such a comparison risks being misleading since EAGERx represents an abstraction based on the actor model \citep{hewitt1973session} and can operate atop a backend like ROS.
This abstraction layer offers vital functionality for robot learning, including synchronized faster-than-real-time simulation, domain randomization, and delay simulation, not inherently supported by ROS.
Recent research \citep{larsen2021reactive} presented a reactive solution to ROS's asynchronous programming challenges via an event-driven API, inspiring EAGERx's synchronization approach. 
However, this API didn't specifically aim to synchronize simulations using expected rates and delays, as demonstrated in our work. 
Importantly, EAGERx's protocol extends beyond ROS to other backends as well.

The proposed synchronization protocol can be seen as an application of the actor model for computation  \citep{hewitt1973session}.
It is a powerful and flexible model of concurrent computation where actors, the primary units, execute tasks concurrently and communicate by exchanging messages. 
The actor model is well-suited for synchronizing robotic systems represented as graphs of nodes, where various nodes need to operate concurrently.
Our protocol operates on an event-driven basis and circumvents dependence on a global/local clock, a central coordinator \citep{raynal2013concurrent}, or extra synchronization messages \citep{messerschmitt1990synchronization}.
Instead, it assesses conditions for subsequent callbacks exclusively after finalizing the preceding one or obtaining a new input channel message.
This can outperform busy-waiting techniques (or spinlock) \citep{anderson1990performance} that continuously evaluate conditions at a fixed time interval.

Ptolemy II \citep{ptolemaeus2014system} constitutes a software framework for designing, modeling, and simulating heterogeneous systems. 
Like EAGERx, it applies the actor model of computation, enabling concurrency and asynchronous communication.
Both frameworks offer graphical user interfaces for visualizing complex systems.
Ptolemy II holds an advantage over EAGERx in its support for a wider range of computation models and the ability to combine them within a single system.
Nevertheless, Ptolemy II serves as a general-purpose framework, while EAGERx specifically targets robot learning.
Furthermore, Ptolemy II employs a Java-based structure, in contrast with EAGERx's exclusive use of Python.

In comparison to Gym \citep{brockman2016openai} — which offers a flexible API but lacks a unified sim2real framework — EAGERx addresses this deficiency.
Unlike Gym's default sequential simulation, EAGERx supports concurrent, distributed operations across devices within environments, enhancing its applicability to robot learning.
Gym environments use object-oriented classes, frequently constructed via inheritance and extended with wrapper patterns.
However, this approach in Gym, particularly with extensive use of wrappers, tends to create overly complex and difficult-to-manage class structures in robotic systems, leading to maintenance challenges and reduced clarity in system design.
Additionally, incorporating time abstraction within Gym environments is challenging, often confining it to multiples of the environment's step size.
Conversely, EAGERx allows each node within the graph environment to operate at separate frequencies.


\begin{table}[t]
\rowcolors{2}{gray!25}{white}
\centering
\resizebox{\linewidth}{!}{%
\begin{tabular}{lccccc}
\hline
                             & EAGERx & \rev{Orbit} \cite{mittal2023orbit} & \rev{Drake} \cite{tedrake2019drake} & \rev{robo-gym} \cite{lucchi2020robo} & \rev{gym-gazebo2}\cite{gonzalez2019gym}\\ \hline
Engine Agnostic              & \cmark & \xmark & \xmark & \pcmark  & \xmark
\\
Specialized Reset Procedures & \cmark & \xmark & \xmark & \xmark & \xmark
\\
Unified Pipeline Sim/Real    & \cmark & \pcmark & \pcmark & \pcmark & \pcmark
\\
Synchronized Simulation      & \cmark & \cmark & \cmark & \xmark & \xmark
\\
Distributed Computing        & \cmark & \cmark & \cmark & \cmark & \cmark \\
GPU Accelerated              & \xmark & \cmark & \xmark & \xmark & \xmark \\
Gradient Information Available & \xmark & \xmark & \cmark & \xmark & \xmark \\
Domain/Delay Randomization & \cmark \color{black}/ \cmark & \cmark \color{black} / \xmark & \cmark \color{black} / \cmark & \xmark \color{black} / \xmark & \xmark \color{black} / \xmark
\\
Environment Visualization    & \cmark & \cmark & \cmark & \pcmark & \pcmark \\
Open Source / License-free                 & \cmark \color{black} / \cmark &  \pcmark \color{black} / \pcmark & \cmark \color{black} / \cmark & \cmark \color{black} / \cmark & \cmark \color{black} / \cmark \\
Documentation / Tutorials    & \cmark \color{black} / \cmark & \cmark \color{black} / \cmark & \cmark \color{black} / \cmark & \pcmark \color{black} / \xmark & \xmark \color{black} / \xmark \\
Last commit (age)  &  $< 1$ week & $2$ months & $< 1$ week & $1$ year & $4$ years \\
\hline
\end{tabular}
  }
\caption{A comparison of various modular sim-to-real robot learning frameworks, where \pcmark \color{black} \hspace{1pt} indicates partial feature presence.
}
\vspace{\vspacebelow}
\label{tab:framework_comparison}
\end{table}

Various robot learning frameworks with connections to EAGERx have been introduced in the field.
Among these, Isaac Orbit \cite{mittal2023orbit} and Drake \cite{tedrake2019drake} stand out as recent frameworks with shared design principles.
In line with EAGERx, Orbit and Drake adopt a modular approach to constructing robot environments, enabling the execution of different nodes at varying rates to support both lower and higher-level control for effective robot learning.
However, these frameworks exhibit three critical differences with EAGERx.
Firstly, EAGERx is designed to be engine-agnostic, whereas Orbit relies on a proprietary simulator, and Drake incorporates an integrated multi-body dynamics simulator, hence restricting them to a single simulation platform.  
Secondly, EAGERx features dedicated reset procedures in the form of reset nodes.
These nodes can be added to the \textit{graph} and are only activated during environment resets.
Thirdly, EAGERx offers a unified pipeline for both simulation and reality.
Although Orbit and Drake promote component reusability in both simulation and reality, EAGERx enforces this more rigorously through \textit{engine-agnostic} and \textit{engine-specific graphs}.
This effectively isolates the engine-agnostic code and minimizes the risk of discrepancies.
Additional frameworks such as Robo-Gym \cite{lucchi2020robo} and Gym-Gazebo(2) \cite{gonzalez2019gym} aimed to exploit the node structure of ROS for robot learning and were primarily centered around the Gazebo simulator without synchronization. 
To speed up training, EAGERx uses multi-processing instead of complete GPU acceleration for parallelization across multiple environments. 
While GPU parallelization can significantly speed up learning \citep{rudin2022learning}, its practicality can sometimes be limited for simulations requiring CPU-bound computations or non-GPU-adaptable libraries. 
In such cases, the latency from data transfer between GPU and CPU can become the dominant factor in simulation speed \citep{freeman2021brax}.
Among the frameworks discussed, only Orbit currently enables parallel training on a GPU. 
A comparative summary of the discussed robot learning frameworks is presented in \tabref{tab:framework_comparison}.

%% file: conclusion.tex
\section{Conclusion}
\label{sec:conclusion}
This paper presented EAGERx, a novel framework to facilitate the transfer of robot learning policies from simulation to the real world.
Our unified framework is compatible with simulated and real robots.
\rev{Its design can accommodate }various abstractions and simulators. 
The presented synchronization protocol simulates delays without sacrificing simulation speed or accuracy, enabling effective policy training in simulation and subsequent transfer to real robots.
We evaluated our framework on two benchmark robotic tasks, demonstrating its effectiveness in reducing the sim2real gap. 
Finally, we demonstrated the utility of the framework beyond sim2real robot learning in two real-world robotic use cases.

\rev{
We plan to extend the open-source code base with more code examples for future work.
    Also, training can be sped up using GPU acceleration, and gradient information can be provided to facilitate optimization through nodes.
    Lastly, it will be valuable to provide real2sim functionalities to reduce the sim2real gap further using real-world data.
}


%% file: root.bbl
\begin{thebibliography}{10}
\providecommand{\url}[1]{#1}
\csname url@samestyle\endcsname
\providecommand{\newblock}{\relax}
\providecommand{\bibinfo}[2]{#2}
\providecommand{\BIBentrySTDinterwordspacing}{\spaceskip=0pt\relax}
\providecommand{\BIBentryALTinterwordstretchfactor}{4}
\providecommand{\BIBentryALTinterwordspacing}{\spaceskip=\fontdimen2\font plus
\BIBentryALTinterwordstretchfactor\fontdimen3\font minus
  \fontdimen4\font\relax}
\providecommand{\BIBforeignlanguage}[2]{{%
\expandafter\ifx\csname l@#1\endcsname\relax
\typeout{** WARNING: IEEEtran.bst: No hyphenation pattern has been}%
\typeout{** loaded for the language `#1'. Using the pattern for}%
\typeout{** the default language instead.}%
\else
\language=\csname l@#1\endcsname
\fi
#2}}
\providecommand{\BIBdecl}{\relax}
\BIBdecl

\bibitem{rudin2022learning}
\BIBentryALTinterwordspacing
N.~Rudin, D.~Hoeller, P.~Reist, and M.~Hutter, ``{Learning to Walk in Minutes
  Using Massively Parallel Deep Reinforcement Learning},'' in \emph{Proc.~of
  the ~Conf.~Robot Learning (CoRL)}, ser. Proceedings of Machine Learning
  Research, vol. 164.\hskip 1em plus 0.5em minus 0.4em\relax PMLR, 08--11 Nov
  2022, pp. 91--100. [Online]. Available:
  \url{https://proceedings.mlr.press/v164/rudin22a.html}
\BIBentrySTDinterwordspacing

\bibitem{tan2018sim}
J.~Tan, T.~Zhang, E.~Coumans, A.~Iscen, Y.~Bai, D.~Hafner, S.~Bohez, and
  V.~Vanhoucke, ``{Sim-to-real: Learning agile locomotion for quadruped
  robots},'' \emph{arXiv preprint}, 2018.

\bibitem{brockman2016openai}
G.~Brockman, V.~Cheung, L.~Pettersson, J.~Schneider, J.~Schulman, J.~Tang, and
  W.~Zaremba, ``{OpenAI Gym},'' \emph{arXiv preprint}, 2016.

\bibitem{hofer2021sim2real}
S.~{Höfer et al.}, ``{Sim2Real in Robotics and Automation: Applications and
  Challenges},'' \emph{IEEE trans.~on Automation Science and Engineering},
  vol.~18, no.~2, pp. 398--400, 2021.

\bibitem{lucchi2020robo}
M.~Lucchi, F.~Zindler, S.~M{\"u}hlbacher-Karrer, and H.~Pichler, ``{robo-gym --
  An Open Source Toolkit for Distributed Deep Reinforcement Learning on Real
  and Simulated Robots},'' in \emph{Proc.~of the IEEE/RSJ Intl.~Conf.~on
  Intelligent Robots and Systems (IROS)}, 2020.

\bibitem{gonzalez2019gym}
\BIBentryALTinterwordspacing
N.~Lopez, Y.~Leire, E.~Nuin, E.~Moral, L.~Usategui, S.~Juan, A.~Rueda,
  M.~Vilches, R.~Kojcev, and A.~Robotics, ``{gym-gazebo2, a toolkit for
  reinforcement learning using ROS 2 and Gazebo},'' \emph{arXiv preprint}, 3
  2019. [Online]. Available: \url{https://arxiv.org/abs/1903.06278v2}
\BIBentrySTDinterwordspacing

\bibitem{quigley2009ros}
M.~Quigley, K.~Conley, B.~Gerkey, J.~Faust, T.~Foote, J.~Leibs, R.~Wheeler, and
  A.~Ng, ``{ROS: an open-source Robot Operating System},'' in \emph{Proc.~of
  the IEEE Intl.~Conf.~on Robotics \& Automation (ICRA)}, vol.~3.\hskip 1em
  plus 0.5em minus 0.4em\relax Kobe, Japan, 2009, p.~5.

\bibitem{koenig2004design}
N.~Koenig and A.~Howard, ``{Design and use paradigms for Gazebo, an open-source
  multi-robot simulator},'' in \emph{Proc.~of the IEEE/RSJ Intl.~Conf.~on
  Intelligent Robots and Systems (IROS)}, vol.~3, 2004, pp. 2149--2154 vol.3.

\bibitem{mittal2023orbit}
M.~{Mittal et al.}, ``{Orbit: A Unified Simulation Framework for Interactive
  Robot Learning Environments},'' \emph{IEEE Robotics and Automation Letters
  (RA-L)}, pp. 1--8, 2023.

\bibitem{tedrake2019drake}
R.~Tedrake and the Drake Development~Team, ``Drake: Model-based design and
  verification for robotics,'' \url{https://drake.mit.edu}, 2019.

\bibitem{nvidia2022isaacsim}
{NVIDIA}, ``{NVIDIA Isaac Sim},'' \url{https://developer.nvidia.com/isaac-sim},
  2022.

\bibitem{coumans2021pybullet}
E.~Coumans and Y.~Bai, ``{PyBullet, a Python module for physics simulation for
  games, robotics and machine learning},'' \url{http://pybullet.org},
  2016--2021.

\bibitem{haarnoja2019soft}
T.~Haarnoja, A.~Zhou, K.~Hartikainen, G.~Tucker, S.~Ha, J.~Tan, V.~Kumar,
  H.~Zhu, A.~Gupta, P.~Abbeel, and S.~Levine, ``{Soft Actor-Critic Algorithms
  and Applications},'' \emph{arXiv preprint}, 2019.

\bibitem{raffin2021stable}
A.~Raffin, A.~Hill, A.~Gleave, A.~Kanervisto, M.~Ernestus, and N.~Dormann,
  ``{Stable-baselines3: Reliable reinforcement learning implementations},''
  \emph{Journal on Machine Learning Research~(JMLR)}, 2021.

\bibitem{andrychowicz2017hindsight}
M.~Andrychowicz, F.~Wolski, A.~Ray, J.~Schneider, R.~Fong, P.~Welinder,
  B.~McGrew, J.~Tobin, P.~Abbeel, OpenAI, and W.~Zaremba, ``Hindsight
  experience replay,'' in \emph{nips}, 2017.

\bibitem{kooi2021inclined}
J.~Kooi and R.~Babu{\v{s}}ka, ``{Inclined Quadrotor Landing using Deep
  Reinforcement Learning},'' in \emph{Proc.~of the IEEE/RSJ Intl.~Conf.~on
  Intelligent Robots and Systems (IROS)}.\hskip 1em plus 0.5em minus
  0.4em\relax IEEE, 2021, pp. 2361--2368.

\bibitem{schulman2017proximal}
J.~Schulman, F.~Wolski, P.~Dhariwal, A.~Radford, and O.~Klimov, ``{Proximal
  policy optimization algorithms},'' \emph{arXiv preprint arXiv:1707.06347},
  2017.

\bibitem{huang2022cleanrl}
S.~Huang, R.~F.~J. Dossa, C.~Ye, J.~Braga, D.~Chakraborty, K.~Mehta, and J.~G.
  Araújo, ``{CleanRL: High-quality Single-file Implementations of Deep
  Reinforcement Learning Algorithms},'' \emph{Journal of Machine Learning
  Research}, vol.~23, no. 274, pp. 1--18, 2022.

\bibitem{todorov2012mujoco}
E.~Todorov, T.~Erez, and Y.~Tassa, ``Mujoco: A physics engine for model-based
  control,'' in \emph{2012 IEEE/RSJ international conference on intelligent
  robots and systems}.\hskip 1em plus 0.5em minus 0.4em\relax IEEE, 2012, pp.
  5026--5033.

\bibitem{shridhar2022cliport}
M.~Shridhar, L.~Manuelli, and D.~Fox, ``{Cliport: What and where pathways for
  robotic manipulation},'' \emph{Proc.~of the ~Conf.~Robot Learning (CoRL)},
  pp. 894--906, 2022.

\bibitem{luijkx2022partnr}
J.~Luijkx, Z.~Ajanovi{\'c}, L.~Ferranti, and J.~Kober, ``{PARTNR: Pick and
  place Ambiguity Resolving by Trustworthy iNteractive leaRning},'' in
  \emph{{5th NeurIPS Robot Learning Workshop: Trustworthy Robotics}}, 2022.

\bibitem{hewitt1973session}
C.~Hewitt, P.~Bishop, and R.~Steiger, ``{Session 8 formalisms for artificial
  intelligence a universal modular actor formalism for artificial
  intelligence},'' in \emph{Advance Papers of the Conference}, vol.~3.\hskip
  1em plus 0.5em minus 0.4em\relax Stanford Research Institute Menlo Park, CA,
  1973, p. 235.

\bibitem{larsen2021reactive}
H.~Larsen, G.~van~der Hoorn, and A.~W{\c a}sowski, \emph{{Reactive Programming
  of Robots with RxROS}}, ser. Studies in Computational Intelligence.\hskip 1em
  plus 0.5em minus 0.4em\relax Springer Verlag, 2021, vol.~6, pp. 55--83.

\bibitem{raynal2013concurrent}
M.~Raynal, \emph{Concurrent programming: algorithms, principles, and
  foundations}.\hskip 1em plus 0.5em minus 0.4em\relax Springer Verlag, 2013.

\bibitem{messerschmitt1990synchronization}
D.~G. Messerschmitt, ``{Synchronization in digital system design},'' \emph{IEEE
  Journal on Selected Areas in Communications}, vol.~8, no.~8, pp. 1404--1419,
  1990.

\bibitem{anderson1990performance}
T.~Anderson, ``{The performance of spin lock alternatives for shared-memory
  multiprocessors},'' \emph{IEEE trans.~on Parallel and Distributed Systems},
  vol.~1, no.~1, pp. 6--16, 1990.

\bibitem{ptolemaeus2014system}
C.~Ptolemaeus, \emph{{System design, modeling, and simulation: using Ptolemy
  II}}.\hskip 1em plus 0.5em minus 0.4em\relax Ptolemy. org Berkeley, 2014,
  vol.~1.

\bibitem{freeman2021brax}
C.~D. {Freeman et al.}, ``{Brax-A Differentiable Physics Engine for Large Scale
  Rigid Body Simulation},'' in \emph{Proc.~of the Advances in Neural
  Information Processing Systems (NeurIPS)}, 2021.

\end{thebibliography}
